\documentclass{article}

\usepackage{microtype}
\usepackage{graphicx}
\usepackage{subcaption}
\usepackage{booktabs} 

\usepackage{hyperref}

\usepackage[preprint]{icml2026}

\usepackage{amsmath}
\usepackage{amssymb}
\usepackage{mathtools}
\usepackage{amsthm}

\usepackage[capitalize,noabbrev]{cleveref}

\theoremstyle{plain}

\theoremstyle{definition}

\theoremstyle{remark}

\usepackage[textsize=tiny]{todonotes}

\newcommand{\method}{\texttt{ErrorMap}}
\newcommand{\product}[0]{\texttt{ErrorAtlas}}

\NewDocumentCommand{\temp}{o}{%
  \textcolor{purple}{[TBD\IfValueT{#1}{ #1}]}%
}

\usepackage{epigraph}
\usepackage{enumitem}
\usepackage{makecell}
\usepackage{multirow}
\usepackage{tabularx}
\usepackage{listings}
\usepackage{wrapfig}

\lstdefinelanguage{json}{
    basicstyle=\ttfamily\small,
    numbers=left,
    numberstyle=\tiny,
    stepnumber=1,
    numbersep=5pt,
    showstringspaces=false,
    breaklines=true,
    frame=single,
    backgroundcolor=\color{gray!10},
}

\icmltitlerunning{\method{} and \product{}: Charting the Failure Landscape of Large Language Models}

\begin{document}

\twocolumn[
\icmltitle{\method{} and \product{}: Charting \\ the Failure Landscape of Large Language Models}

\icmlsetsymbol{equal}{*}

\begin{icmlauthorlist}
    \icmlauthor{Shir Ashury-Tahan}{ibm}
    \icmlauthor{Yifan Mai}{stanford}
    \icmlauthor{Elron Bandel}{ibm}
    \icmlauthor{Michal Shmueli-Scheuer}{ibm,equal}
    \icmlauthor{Leshem Choshen}{ibm,mit,equal}
\end{icmlauthorlist}

\icmlaffiliation{ibm}{IBM Research}
\icmlaffiliation{stanford}{Stanford University}
\icmlaffiliation{mit}{MIT}




\icmlcorrespondingauthor{Shir Ashury-Tahan}{shir.ashury.tahan@ibm.com}
\icmlcorrespondingauthor{Leshem Choshen}{leshem.choshen@ibm.com}

\icmlkeywords{Machine Learning, ICML}

\vskip 0.3in
]
\printAffiliationsAndNotice{\textsuperscript{*}\;Equal supervision.}



\begin{abstract}

Large Language Models (LLM) benchmarks tell us when models fail, but not \textit{why} they fail. A wrong answer on a reasoning dataset may stem from formatting issues, calculation errors, or dataset noise rather than weak reasoning. Without disentangling such causes, benchmarks remain incomplete and cannot reliably guide model improvement.
We introduce \method{}, the first method to chart the sources of LLM failure. It extracts a model's unique ``failure signature'', clarifies what benchmarks measure, and broadens error identification to reduce blind spots. 
This helps developers debug models, aligns benchmark goals with outcomes, and supports informed model selection. \method{} works on any model or dataset with the same logic.
Applying our method to $35$ datasets and $83$ models we generate \product{}, a taxonomy of model errors, revealing recurring failure patterns. \product{} highlights error types that are currently underexplored in LLM research, such as omissions of required details in the output and question misinterpretation.
By shifting focus from where models succeed to why they fail, \method{} and \product{} enable advanced evaluation --- one that exposes hidden weaknesses and directs progress. 
Unlike success, typically measured by task-level metrics, our approach introduces a deeper evaluation layer that can be applied globally across models and tasks, offering richer insights into model behavior and limitations.
We make the taxonomy and code publicly available\footnote{\url{https://github.com/IBM/ErrorMap}},
with plans to periodically update \product{} as new benchmarks and models emerge.

\end{abstract} 

\vspace{-1cm}
\begin{flushleft}
\epigraph{
``\textit{It is possible to fail in many ways\ldots{}\\ while to succeed is possible only in one way.}''}
{\textit{Aristotle, \emph{Nicomachean Ethics}, Book~II, $\sim$320BC}}
\end{flushleft}
\vspace{-0.4cm}

\begin{figure*}
    \centering
    \includegraphics[width=\linewidth]{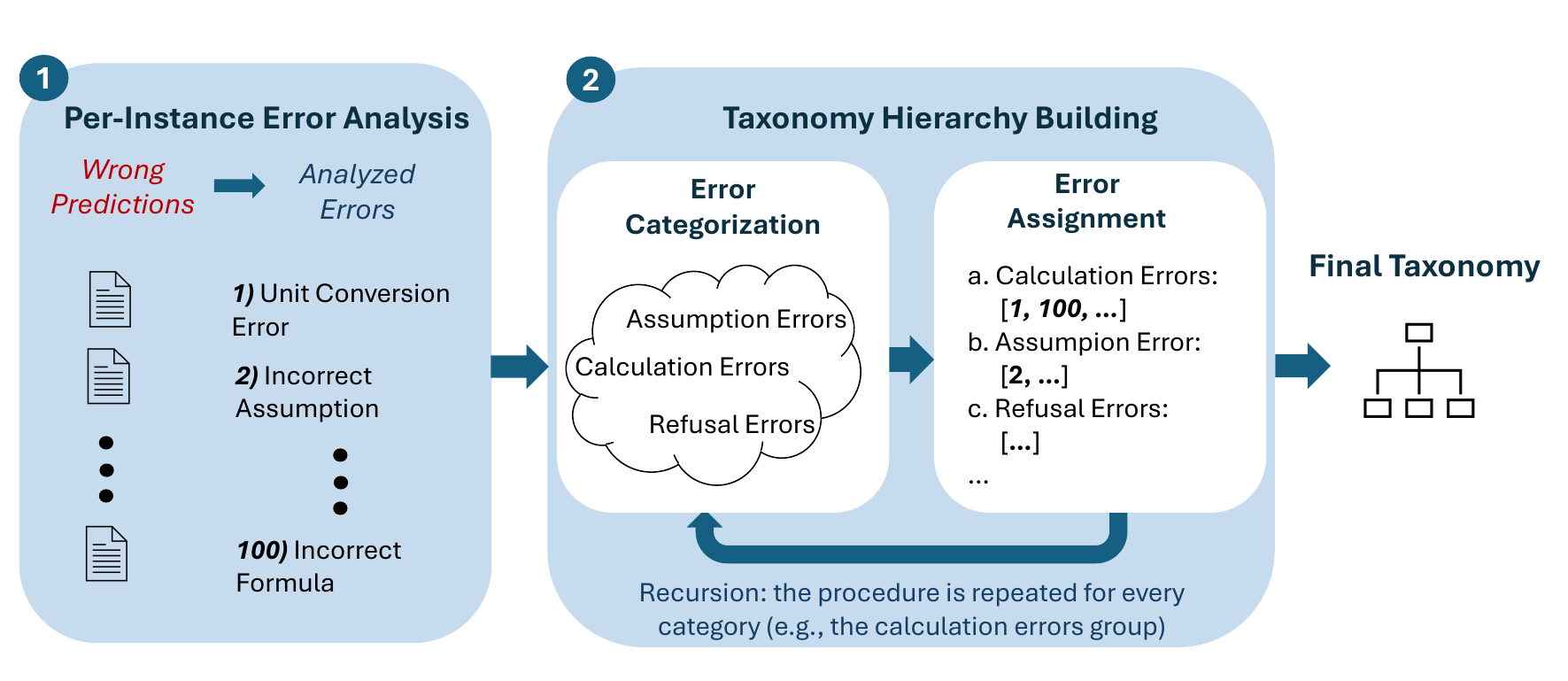}
    \caption{ErrorMap Schema: Two Method Stages. The first stage performs a one-time analysis of each incorrect prediction at the instance level. The second stage constructs a layered taxonomy through recursive iterations, based on either the maximum grouping depth or a defined layer threshold.}
    \label{fig:error_map_schema}
\end{figure*}

\section{Introduction}

Benchmarking plays a central role in advancing large language models (LLMs), offering a standard bottom-line score to assert progress \citep{Biderman2024LessonsFT}. This abstraction eases proving a model's success or its overall superiority, but it also obscures the nature and origin of model errors, complicating skill comparisons and hindering efforts to diagnose limitations or guide improvements.

In response to these limitations, there is growing interest in developing more interpretable and diagnostic evaluation frameworks \citep{maimon2025iqtestllmsevaluation, zeng2025evaltreeprofilinglanguagemodel, tjuatja2025behaviorboxautomateddiscoveryfinegrained} or in highlighting specific errors models make \citep{mukherjee2025premiseaugmentedreasoningchainsimprove, Pan2025LEMMALF, Li2024EvaluatingMR, Honovich2022TRUERF, Kryscinski2019EvaluatingTF}. While current diagnostic methods offer valuable insights, their analysis primarily relies on the challenges posed by the input (e.g., counting failures on differential equations questions). However, where success on a challenge necessarily proves competence, a failure can have many causes (e.g., misunderstanding the question, miscalculation or applying a wrong axiom). Moreover, benchmark examples themselves may introduce ambiguity or error, further complicating evaluation, especially when the model's answer is not considered. Ultimately, pinpointing the cause of failure requires analyzing both the input, including the question and the instruction, and the resulting answer.


We introduce \method{} (\S\ref{sec:method}) to address this gap. \method{} offers a model-oriented error analysis, highlighting why models fail. The method transforms raw language model failures into a structured, interpretable taxonomy in natural language. To do so, \method{} follows a pipeline, first profiling the issue underlying each failure, then recursively generates taxonomy categories through an iterative refinement process, and finally applies the constructed taxonomy to all failures (see Figure~\ref{fig:error_map_schema}).

As our analysis requires simple unstructured text, it applies seamlessly to any language model and domain. Overall, \method{} provides a flexible way to analyze a practitioner's specific setting or to compare the failure fingerprint of several models or datasets. While a dynamic taxonomy that would fit an ad hoc analysis is best for many needs, a stable taxonomy simplifies comparisons across time and replicability.

Applying \method{}{} to $83$ models and $35$ datasets, we release \product{} -- a taxonomy of current LLM failures (see 
\S\ref{sec:atlas}). \product{} is a comprehensive and static taxonomy of model errors designed to facilitate cross-field comparisons, enhance efficiency, and ensure replicability.

We present several findings on common model failures, which both stand on their own and highlight the effectiveness of our methods.
Applying the \product{} taxonomy to current models and datasets we find issues that are prevalent but understudied (\S\ref{sec:results}). For example, models often misinterpret the question's intent or provide incomplete answers.

Beyond \product{}{}, \method{} can support practitioners throughout the LLM lifecycle, from development and fine-tuning to evaluation and benchmarking. We provide in \S\ref{sec:applicability} two test cases, a model developer diagnosing the differences between two versions of Gemini~\citep{team2024gemini} and a benchmark curator testing MMLU-pro \citep{wang2024mmlu}.

Furthermore, in \S\ref{sec:validate}, we validate that the stages involved in extracting the error analysis are accurate, robust and cover well the errors models make.

This work advances our ability to understand and analyze model behavior at a deeper level. Our contributions are as follows:

\begin{enumerate}
    \item We introduce \method{}, an LLM-based technique to generate a dedicated taxonomy of LLM errors.
    It enables analysis across a diverse set of domains, input formats and model comparisons.
    \item We present \product{}, a static taxonomy of model errors, generated using \method{}. It captures common failure modes across benchmarks and models. These errors reflect underlying limitations in model behavior, supporting meaningful and interpretable comparisons of model weaknesses.
    \item We provide analysis across a large number of models and datasets. Finding common errors that are understudied. Moreover, we find model versions, types, and families exhibit distinct error patterns, allowing for nuanced behavioral profiling and more targeted evaluation.
    \item We demonstrate the applicability of both \method{} and \product{} for nuanced model comparison, benchmark analysis, and model debugging.
    \item We publicly release the code, taxonomy and associated data.
\end{enumerate}

\section{\method{}} \label{sec:method}

Our technique targets a common scenario: evaluating multiple models on the same dataset examples, as typically done in benchmark runs, or even reusing a benchmark run for deeper analysis. It leverages all available data in the benchmark, including inputs, reference answers, and model outputs, and produces comparative insights across models.
The process is unsupervised and consists of two stages. First, incorrect predictions are analyzed on a per-example basis to form the leaf nodes of the taxonomy, creating the initial layer. Second, the taxonomy is constructed top-down through recursive process. 
The method schema overview is shown in Figure~\ref{fig:error_map_schema}, and additional details, including the specific prompts used, are provided in Appendix~\ref{error_map_additional_details}.

\textbf{Stage 1: Per-Instance Error Analysis}
Our goal at this stage is to create a structured summary of the resulted errors and provide interpretable analysis of it.
To achieve this, we task an analyst LLM with performing a detailed, structured analysis for each incorrect prediction. This includes evaluating a list of criteria with associated features, providing a summary of the failure, and assigning a short \emph{label} for it.
To support the judge’s evaluation, we provide the following information: the original instance, any available references and multiple Informative Correct Predictions (ICPs) if available, i.e., correct predictions
made by other models in the benchmark. 
ICPs have proven useful  \citep{zelikman2022starbootstrappingreasoningreasoning, LI2022ExplanationsFL, Creswell2022FaithfulRU}. In this context, they act as rich reference points that often approximate full solutions, helping judges compare correct and incorrect outputs rather than diagnose root causes. This is particularly valuable when no gold reference exists or when the gold standard is limited to a final answer (e.g., in classification tasks).


The judge is asked to construct a structured solution to the instance (see prompt in Appendix~\ref{app:per_instance_prompt}.) The structure has several components, all of which the judge should fill. The judge is asked to break down the solution and specify \emph{criteria}; steps, evidence, or assumptions required to reach a correct answer, for example relying on formulas, a list of reasoning steps of extracting multiple facts to deduce and answer. For each criterion, the judge should assess its presence, quality, supporting evidence (a quote from the prediction), and may add comments if there is something additional to note about this criterion. Grounded in the step-by-step analysis, the judge should identify the first major error that caused the prediction to fail and create both a \emph{summary} of a few sentences and an informative \emph{label} that highlights the failed skill. We focus on the first major error because it often sets the trajectory for the rest of the reasoning; once an initial mistake is made, subsequent steps are likely to be flawed as well. This label is a phrasal description of the identified error.
Finally, the judge outputs a JSON object that includes the necessary detailed criteria, along with the error summary and label. Note that, while only the error label is used in the next stage, the criteria and summary are helpful for interpreting each specific wrong prediction.\looseness=-1

\textbf{Stage 2: Taxonomy Hierarchy Building} 
This stage organizes the errors analyzed in the previous step into a multi-level taxonomy that enables systematic navigation across different resolution paths. The hierarchy is constructed in a top-down manner through iterative cycles of two substages: (a) extracting categories from the analyzed errors (Stage 2.a) and (b) classifying errors into these categories (Stage 2.b). The initial iteration considers all analyzed errors; thereafter, for each category produced in the preceding Stage 2.a, child categories are derived based on the errors assigned to it in the corresponding Stage 2.b. This process can continue until no further meaningful subdivisions can be identified, but it may also be stopped earlier.

\textit{Substage 2.a: Error Categorization}
This stage consolidates all or selected instance-level results from the previous analysis into a set of common error types by iteratively grouping unique error labels into broader categories. Each category is assigned a clear description to minimize ambiguity.

To construct the categories, we adopted the  data mining approach proposed by \citet{Wan2024TnTLLMTM}, which iteratively employs an LLM to generate categories from input data, in our case, from the unique error labels and their prevalence. We summarize its 3 stages below \citep[c.f.,][]{Wan2024TnTLLMTM}, and provide in Appendix their prompts (\S\ref{app:taxonomy_generation_prompt}, \S\ref{app:taxonomy_update_prompt}, \S\ref{app:taxonomy_review_prompt}) and configuration (Table~\ref{tab:taxonomy_config_params}).

\begin{enumerate}[itemsep=0.5pt, topsep=0pt,leftmargin=1.4em,labelsep=0.5em]
    \item \textit{Category Generation} – The initial stage, where the LLM receives the first batch (a list of error labels with their frequencies) and generates categories and category descriptions based on it. Note that since error labels from stage 1 were created in free-form, label repetitions were not guaranteed, though we observed frequent overlaps.
    
    \item \textit{Iterative Refinement} – Multiple iterations (depending on data size), where the LLM receives the previously generated categories along with a new sampled batch and incrementally updates and improves the categories. 
    
    \item \textit{Final Review} – A concluding iteration where the LLM reviews the final taxonomy to ensure coherence and compliance with the instructions (e.g., no ambiguity).
\end{enumerate}

The output of this stage is a finalized list of categories, each accompanied by a clear description, derived from the corresponding group of instance-level errors.

\textit{Substage 2.b: Error Assignment}
This stage integrates the instance-level analysis performed in the previous step and assigns each error, along with its criteria analysis, summary, and label to the most appropriate category, based on the classification of its error label.

This integration is done using a simple batched LLM call (the prompt is provided in App. \S\ref{app:classify_prompt}). We provide the model with the error categories and a batch of error labels (see parameters in Appendix, Table~\ref{tab:taxonomy_config_params}), and ask the model to assign each error to the most appropriate category. 

The outcome of this stage is a set of categories, each populated with instance-level errors relevant to it.


\textit{Final Outcome}
The final outcome of this stage is a layered taxonomy, where multiple category levels each contain a list of errors grouped by their respective labels.

\begin{table*}[ht]
\centering
\footnotesize
\begin{tabular}{p{4cm} p{12cm}}  

\toprule
Category & Description \\
\midrule
Logical Reasoning Error & Fails in logical inference, deduction, or applying correct reasoning steps. \\
Missing Required Element & Omits mandatory sections, fields, identifiers, or other specified content. \\
Computation Error & Produces incorrect numerical, algebraic, or geometric results, including miscalculations and faulty derivations. \\
Incorrect Identification & Mislabels or misidentifies objects, concepts, attributes, or entities. \\
Specification Misinterpretation & Misunderstands task requirements, output type, or provides incorrectly formatted parameters and inputs. \\
Output Formatting Error & Violates required structure, markup, punctuation, case, or other formatting rules. \\
Irrelevant/Extraneous Content & Generates off‑topic, unrelated, or unnecessary information beyond the required answer. \\
Counting/Enumeration Error & Over‑counts, under‑counts, double‑counts, or omits cases in combinatorial reasoning. \\
Answer Selection Error & Chooses wrong answer option, label, or maps solution to an incorrect choice. \\
Incomplete Reasoning & Lacks essential explanation, proof steps, or logical justification. \\
Factual Error & Provides inaccurate or fabricated factual information or domain knowledge. \\
Tool/API Usage Error & Missing, wrong, or extraneous tool/API calls; misuse of tool capabilities. \\
Naming/Symbol Error & Incorrect function/variable names, symbols, or identifier usage. \\
Inappropriate Refusal & Unjustified or overly restrictive refusal to answer the query. \\
Unit Conversion Error & Incorrect conversion between units, percentages, temperature, or similar measures. \\
False Positive Detection & Incorrectly flags an error or anomaly that does not actually exist. \\
Error Detection Failure & Fails to recognize existing mistakes, leaving errors undetected. \\
\bottomrule
\end{tabular}
\caption{\product{}: High-level error categories and category descriptions. The categories are sorted most to least prevalent.}
\label{tab:error_atlas_categories}
\vspace{-0.7cm}
\end{table*}

\section{Constructing \product{}}\label{sec:atlas}

Where \method{} supplies a flexible way to acquire a dedicated taxonomy for a nuanced issue, such as specific models or task data, a static taxonomy is often preferred in cases where the replicability and broad comparisons are required. \product{} is built to accommodate such error analysis use-cases. We describe \product{}, a taxonomy that categorizes failure modes commonly shown by current popular models, built using \texttt{gpt-oss-120b}. We detail the process of constructing \product{}, including practical decisions made, such as the identity of the datasets. We refer to experimental details that are general to all our experiments, from building the taxonomy to validating it in Appendix~\ref{sec:exp_setup}.

\textbf{Coverage} To create \product{}, we select a diverse group of $35$ datasets spanning a wide range of tasks, domains, and capabilities, and extract an \method{} taxonomy across all available model predictions. In total, we include predictions from $83$ models.
We cover the scope of LLM evaluations with the following benchmarks: from HELM leaderboards \citep{liang2023holisticevaluationlanguagemodels}, Capabilities for general capabilities \citep{capabilities2024}, MedHELM for medical domain \citep{bedi2025medhelmholisticevaluationlarge}, and ToRR for tables \citep{ashurytahan2025mightytorrbenchmarktable}, and BFCL-v4 \citep{patil2025bfcl} for function calling, and for code HumanEval \citep{chen2021evaluatinglargelanguagemodels}, HumanEval Plus  \citep{liu2023codegeneratedchatgptreally}, MBPP \citep{austin2021program} and MBPP Plus \citep{liu2023codegeneratedchatgptreally}. For both ToRR and MedHELM, we selected partial subsets of the datasets they contain.\footnote{We focused on tasks where model outputs include interpretable content, as \method{} goal is to analyze predictions that reveal failure modes. Many benchmark tasks, like classification or entity extraction do not provide explanations (or CoT) and lack the necessary context for such analysis. The full details of selected datasets are provided in App. Table~\ref{tab:selected_datasets}.}

\textbf{Scaling \method{} through Sampling} The flexible nature of \method{} allows it to be applied to any number of models, datasets, and incorrect predictions. However, constructing \product{}, a unified taxonomy across all mentioned benchmarks, is far more demanding than applying \method{} in narrow settings (e.g., a single benchmark or a small set of models). 
The number of errors to analyze scales with the dataset sizes and number of models, resulting in significant computational demands, a common obstacle for large scale evaluations \citep[][]{perlitz-etal-2024-efficient}. 
To manage the data volume, we employed relative sampling: for each model-dataset pair, we sampled approximately $10\%$ of the instances where the model was evaluated as having failed, i.e., a proportionate subset based on the model’s error rate. This resulted in a sample of approximately $7{,}000$ failures that was then used to run \method{}. 
Interestingly, despite the limited sample size and its variability, the categories in the resulting taxonomy were unique, and no overlaps were observed. This may suggest that the iterative refinement effectively consolidates similar error types and is not sensitive to sample size.


\textbf{Usage} Now that we have extracted \product{}, its primary value is in clearly surfacing common LLM error types (see \S\ref{sec:results}). This can support future model development and real-world improvements, particularly as we uncover previously unreported error categories.
Moreover, \product{} categories can be practically applied at low cost to reflect general model failure modes. This can be done by running only Stage 1 and Stage 2.b, while skipping Stage 2.a (Error Categorization).\footnote{Running stage 2.a on specific data may be less representative from a model's general failure mode perspective, as it depends on data collection that may be biased.}



\section{\product{} Applicability} \label{sec:results}

\textbf{\product{} Reveals the Error Topography of Models}
Running \method{} on $35$ datasets results in the construction of \product{} (see \S\ref{sec:atlas}). The main outcome is a set of $17$ high-level taxonomy categories describing common model errors, presented in Table~\ref{tab:error_atlas_categories}. \product{} statistics are available in App. Table~\ref{tab:error_atals_numbers}, and examples of error categories with their subcategories can be found in App. Table~\ref{tab:taxonomy_hierarchy_samples_subcategories}.

The resulting error categories span a wide spectrum, reflecting diverse dimensions of model performance. These include reasoning-related errors, such as Logical Reasoning Errors, Incorrect Identification, Specification Misinterpretation; instruction-following issues, including Missing Required Element, Irrelevant/Extraneous Content; procedural errors, such as Computation Error, Counting/Enumeration Error, Unit Conversion Error; and technical and linguistic issues, including Output Formatting Error, Inappropriate Refusal, Tool/API Usage Error. Additionally, there are categories that fall outside these dimensions, such as False Positive Detection.

While the areas of failure described above (e.g., instruction following and reasoning) are generally well-known and researched within the community, \product{} enables the identification of more precise weaknesses within these broader categories. Moreover, the resulted taxonomy underscores a key limitation of benchmark scores: although benchmarks provide useful indicators of model performance on specific tasks or domains, they often lack the granularity required to uncover detailed failure patterns. Understanding these patterns is essential for diagnosing concrete limitations in model behavior and guiding targeted improvements.



\begin{figure}
\vspace{-.2cm}
    \centering
    \includegraphics[width=0.51\textwidth]{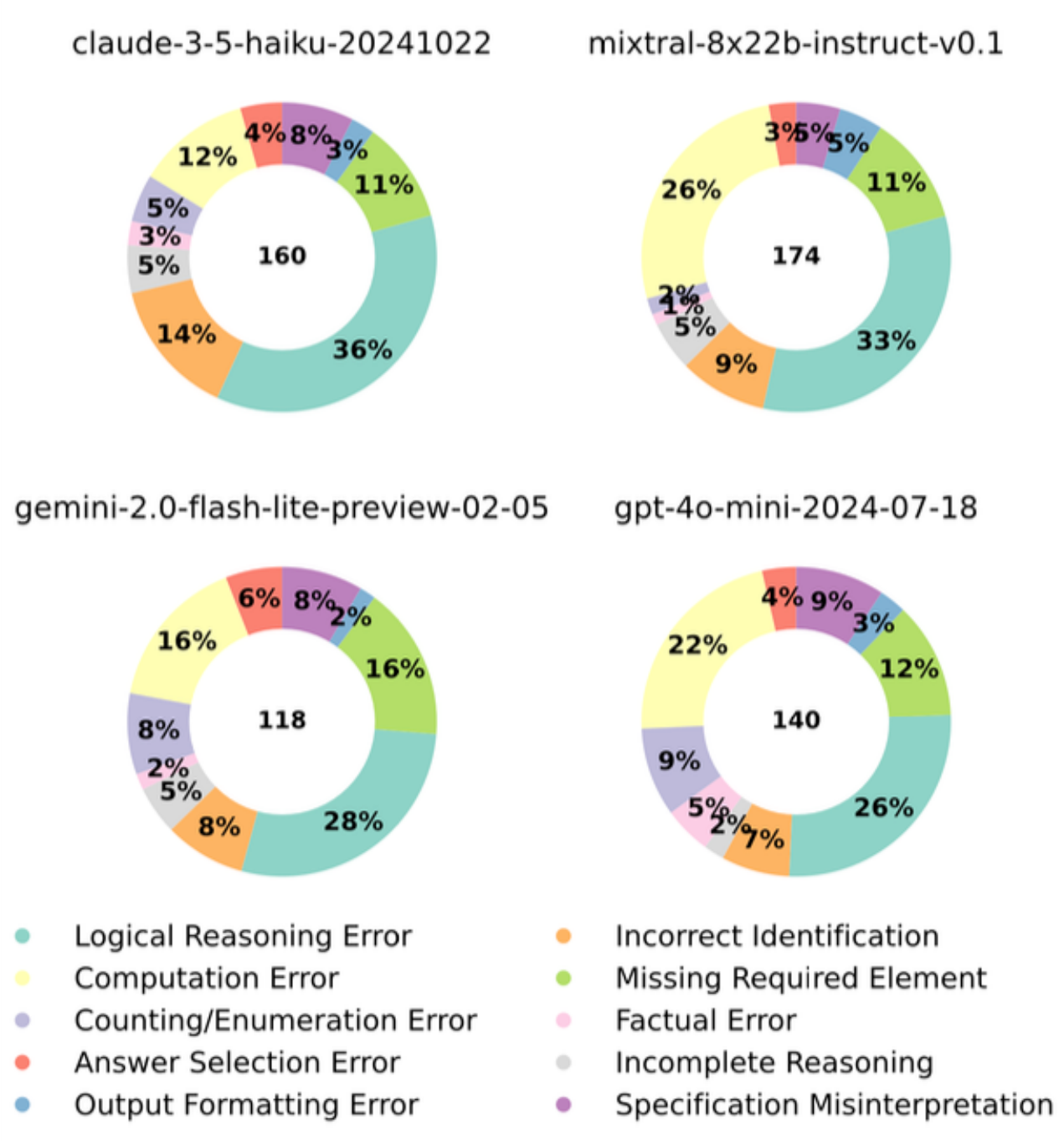}
    \caption{Error distributions for models across the $10$ most prevalent \product{} error categories on the Capabilities benchmark. The total number of errors for each model appears at the center of its corresponding donuts chart. Each error count represents roughly $10\%$ of the model's total errors on Capabilities.}
    \label{fig:model_dist_capabilities}
\vspace{-1cm}
\end{figure}

\begin{table*}[]
\footnotesize
\centering
\begin{tabular}{p{2.6cm} p{2.6cm} p{4cm} p{6cm}}
\toprule
\textbf{Use Case} & \textbf{Persona} & \textbf{Goal} & \textbf{Example} \\
\midrule
Model Debugging & Model Developer & \parbox{5cm}{Identify regressions\\ and behavioral changes} & Compare model versions (e.g., v1 vs. v2) to detect reductions in specific error types, such as reasoning failures, especially when targeting improvements in those areas \\[1.2ex]

Benchmark Analysis & Benchmark Creator & \mbox{Reveal model capabilities}, provide error \mbox{distributions} and debug dataset validity & Run \method{} on benchmark results to characterize model error distribution and error types across tasks \\[1.2ex]

Model Selection & Product Team & Choose the most suitable model for deployment & Select the best model on domain-specific tasks based on stakeholder preferences (e.g., prioritizing fewer hallucinations in medical applications). \\[1.2ex]

Domain-Specific Evaluation & \parbox{5cm}{Domain Expert /\\ Analyst} & Identify failure modes in specialized contexts & Use \method{} to analyze model responses in high-stakes domains (e.g., legal, medical) and surface common failure patterns \\[1.2ex]
\bottomrule
\end{tabular}
\caption{Summary of key use cases using \method{}.}
\label{tab:usecases}
\vspace{-0.7cm}
\end{table*}

\textbf{Surfacing Frequent but Overlooked Model Failures}
While some error types are more actively studied, such as reasoning errors \citep{zheng-etal-2025-processbench,xu2025large,liu2023exposing} and hallucinations \citep{cattan2025localizingfactualinconsistenciesattributable, zhao2024wildhallucinationsevaluatinglongformfactuality} and others can be mitigated through techniques like tool use (e.g., resolving calculation mistakes), \product{} reveals additional error patterns that have received limited attention in the community, despite their prevalence not justifying such disproportionate neglect.

One such pattern, is labeled in \product{} as \textit{Missing Required Element}. Surprisingly, despite being a prevalent error in our analysis, this error type is under-discussed\footnote{Similar issues were hardly mentioned in related work search and existing taxonomies, with the exception or retrieval literature.}. Upon manual inspection of the results, we found that this pattern usually involves missing details with respect to the context, such as not fully answering the question, omitting specific nuances requested, or ignoring certain instructions and constraints. Examples provided in Appendix~\ref{appendix:incomplete_response_examples} show such cases where the model produces a partially correct solution.
This tendency to overlook contextual cues can significantly impact the reliability of AI systems. For instance, while a set of symptoms may typically suggest a particular diagnosis, subtle nuances in a specific case could point to a completely different one.

Another error shown in Tables~\ref{tab:error_atlas_categories},\ref{tab:error_atals_numbers}, with notable prevalence across datasets, is \textit{Specification Misinterpretation}. Examining specific instances in this category reveals various cases where models fail to adequately consider context or respond with the required expertise. These include instances of misalignment between surface cues and deeper context, as well as failures to interpret the information provided in the context (see examples in Appendix~\ref{appendix:ques_misinterpret_examples}).
This underscores the need for improved contextual understanding in model development, particularly for nuance‑sensitive tasks.

\textbf{Error Patterns Vary Between Models.}

Employing \product{}, we observe distinct model-specific patterns that reveal nuanced variations in error behavior.
To quantify these differences, we analyzed the error distributions of models. To ensure a fair comparison, we selected models that appear in the same benchmark, HELM Capabilities, and measured their error distributions within it.
Figure~\ref{fig:model_dist_capabilities} illustrates the top $10$ error categories across several models. Notably, all models tend to make errors related to incomplete content, with \texttt{Gemini 2.0 Flash Lite} exhibiting the highest frequency in this category, while showing the fewest errors in incorrect formatting.
\texttt{Claude 3.5 Haiku} shows a higher tendency toward logical reasoning errros compared to others, whereas \texttt{Mixtral 8x22b Instruct v0.1} is more prone to computation errors.




\begin{figure}

  \centering
  \includegraphics[width=\linewidth]{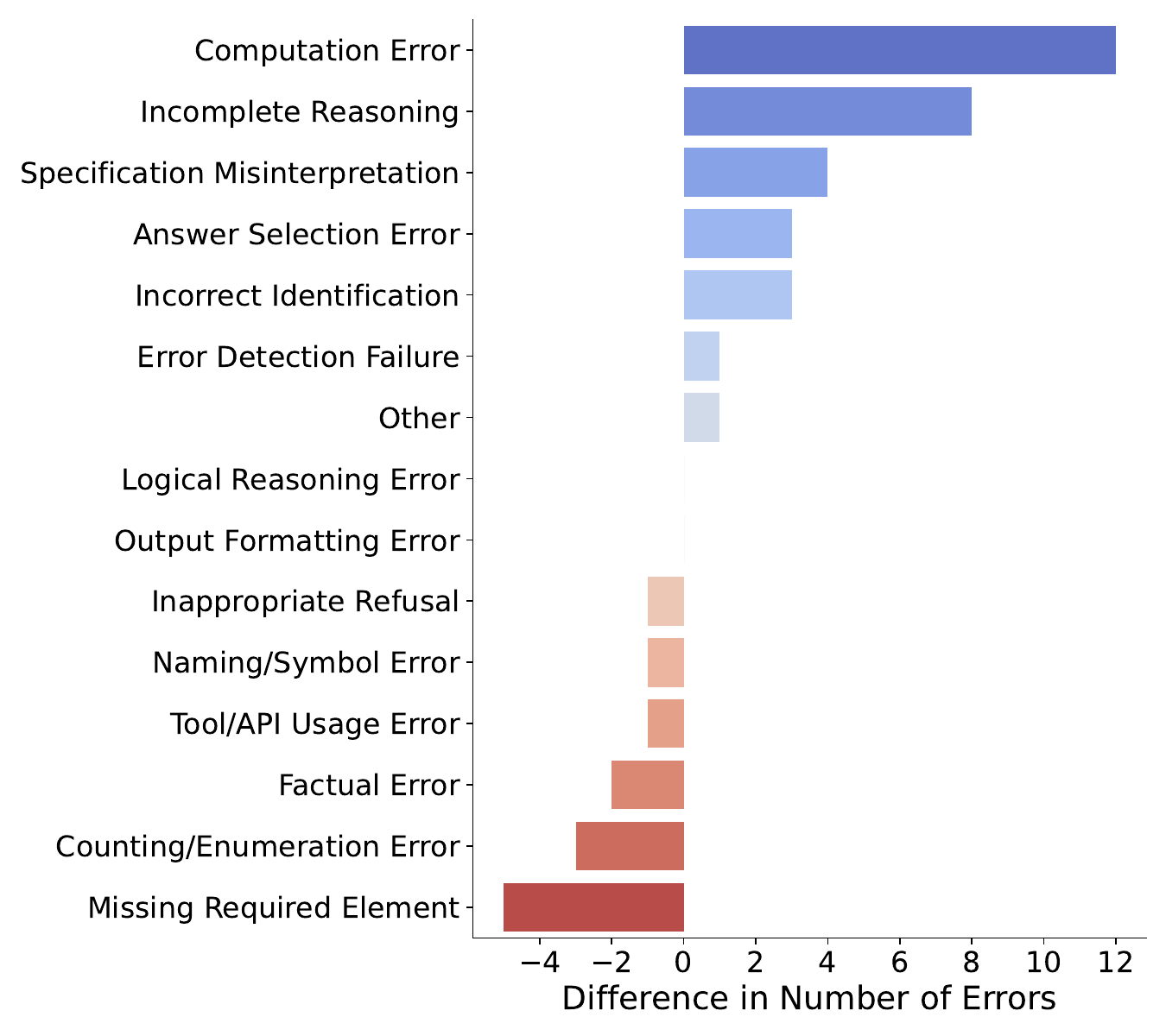}
    \caption{Differences in number of errors per category between \texttt{Gemini 1.5}  flash and Pro on the capabilities benchmark in HELM. X-axis represents the change in error frequency, highlighting areas of improvement or regression.}

    \label{fig:gemini_diff_bars}
    \vspace{-0.5cm}
\end{figure}

\textbf{Model Error Does not Always Reflect Failure in the Benchmark Targeted Skill.}
We motivated our reliance on model outputs by the discrepancy between what a question aims to test and what eventually trips the model. We indeed find such cases in our analysis.
For example, consider capability-focused datasets like MMLU-Pro, Omni-MATH, and GPQA, which are considered challenging due to their reasoning demands. While Omni-MATH emphasizes math reasoning, GPQA focuses on general reasoning, and MMLU-Pro primarily tests knowledge along with a reasoning depth. However, approximately $44\%$ of model errors in these benchmarks have a weak reasoning orientation, and seem more technical challenges, e.g., computation error, missing required element, or counting error  (see App. Table~\ref{tab:mmlu_omni_gpqa_stats}). 

Overall, we have shown the usefulness of \product{} for comparing model providers, models, and gaining insight in a new domain. In \S\ref{sec:applicability}, we discuss the cases where a dedicated taxonomy is helpful and showcase it.


\section{\method{} Applicability} \label{sec:applicability}

In Section~\ref{sec:results} we saw \product{} enables evaluations that help track model weaknesses and monitor progress over time when rerun. However, many diagnostic users are interested in their specific failure modes rather than general ones. Table~\ref{tab:usecases} summarizes five key use cases, each illustrating how \method{} can aid decision-making, debugging, and evaluation in various contexts. In this section we demonstrate the first and second use-cases,
aimed at supporting model developers and benchmark persona. The experimental setup is provided in App. \ref{sec:exp_setup}.

\subsection{Model Developers} \label{subsec:model_developer}

For model developers, \method{} provides a structured way to assess behavioral changes between iterations \citep[similar to a behavioral model-diff;][]{mishra-sharma2025insights,lindsey2024sparsecrosscoders,aranguri2025modeldiff}. For example, it can help answer questions like: What common errors in my setting did my improved version address? or Did integration with external tools reduce hallucinations? By surfacing such differences, \method{} supports more informed and targeted improvements.




To test this, we compared \texttt{Gemini 1.5 flash} and \texttt{Gemini 1.5 pro} using capabilities benchmark data from HELM. While the pro version outperforms the flash version by a mean score of $4.8\%$ on the benchmark, one may wonder what are the differences between them. Our analysis (shown in Figure~\ref{fig:gemini_diff_bars}) presents the differences in number of errors. It is evident that the pro model performs significantly less computation and incomplete reasoning errors. A model developer can use such analysis to determine whether the changes made in the pro version were focused on that aspect, or are those unexpected changes that call for more developmental efforts.

We also note that this analysis can be valuable for other model stakeholders, helping them \textit{make more informed decisions}. For example, if one has the budget to utilize Gemini models but seeks to optimize costs, this evaluation can serve as a valuable guide. In scenarios where the task involves calculations the pro version may be preferable due to its enhanced capabilities. Conversely, for tasks primarily focused on generation, where such precision is less critical, paying more for the pro version might not be cost-effective.

\begin{table}[]
\centering
\resizebox{\columnwidth}{!}{
\begin{tabular}{p{4.5cm} p{4.5cm}}
\toprule
\textbf{\method{} Categories} & \textbf{MMLU-Pro Paper Categories} \\
\midrule
Logical Reasoning Error (44\%) & Reasoning Errors (39\%) \\
Mathematical Mistake (24\%) & Calculation Errors (12\%) \\
\multirow{2}{*}{\begin{tabular}[c]{@{}l@{}}Incomplete Answer (13\%)\\ Factual Error (12\%)\end{tabular}} & \multirow{2}{*}{Lack of Specific Knowledge (35\%)} \\
 & \\
Prompt Misinterpretation (5\%) & Question Understanding Errors (4\%) \\
 & Other (10\%) \\
\bottomrule
\end{tabular}
}
\caption{Comparison of MMLU-Pro error categories and GPT-4o distribution between manual annotations from the original paper and our method.}
\label{tab:mmlu_gpt4o}
\vspace{-1cm}
\end{table}

\subsection{Benchmark Creators}
Diagnosis is also important for benchmark curators to validate what key challenges it poses for models and highlight unexpected errors.
To demonstrate usability, 
we use MMLU-Pro as a case study and apply \method{} to generate its taxonomy, demonstrating several key capabilities of our approach:

\textbf{\method{} closely approximates manual analysis in MMLU-Pro.}
Running \method{} on MMLU-Pro dataset provided us with $5$ error categories. We compared the manual analysis the paper
reported for \texttt{GPT-4o} with ours and got a similar error distribution (see Table~\ref{tab:mmlu_gpt4o}), with the exception of two categories in \method{} that map to one manual one and no ``other'' category.


\textbf{Comparing Dataset Parts for Richer Interpretive Insights}
While \method{} separates analysis bottom up by the errors, integrating high-level data dimensions can yield more nuanced results. To demonstrate this, we analyzed model errors in relation to the domain categories of the dataset, as shown in App. Figure~\ref{fig:mmlu_cat_vs_domains}. Some patterns appear intuitive, for example, mathematics and physics exhibit similar error distributions. However, other findings are less expected, such as the disproportionately high number of factual errors in the health domain, even exceeding those in history. 

\begin{figure}[]
\vspace{-0.3cm}
    \centering
    \includegraphics[width=\linewidth]{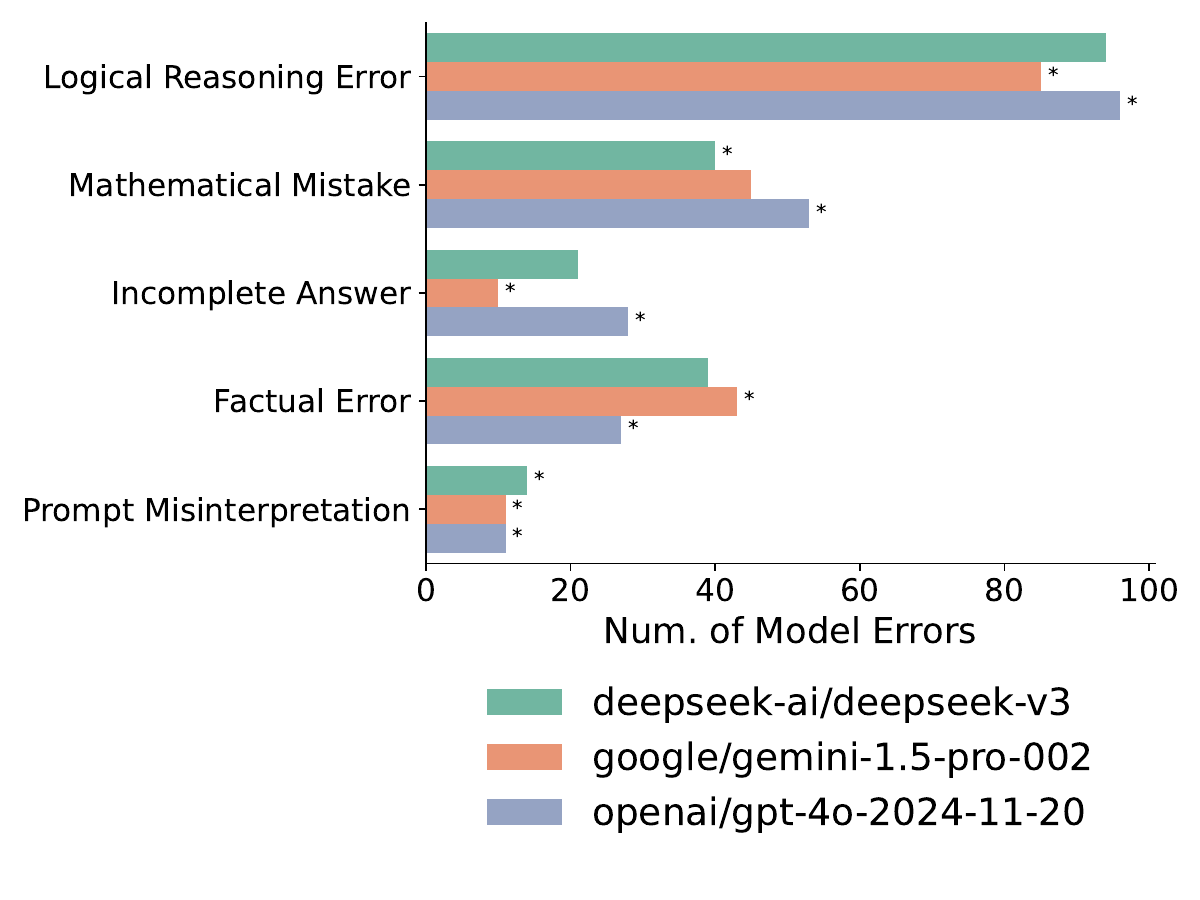}
    \caption{Differences in error category distributions among three leading models on the MMLU-Pro dataset. Asterisks (*) indicate bars compared in statistical significance test (see App. Table~\ref{tab:significance_best_worst})}
    \label{fig:diff_model_dists_mmlu}
    \vspace{-1.2cm}

\end{figure}

We further note that benchmark users can benefit from the model comparisons presented in benchmarks to distinguish between models in a more granular fashion.
To demonstrate this, we compare three models on their MMLU-Pro error distributions in Figure~\ref{fig:diff_model_dists_mmlu}. For instance, we find that \texttt{GPT-4o} exhibits a higher proportion of reasoning errors compared to \texttt{Gemini 1.5 Pro}, which, in contrast, makes significantly more factual errors.

In conclusion, \method{} enhances understanding of benchmark datasets beyond overall metrics. By revealing task-specific insights and taxonomy, it helps users interpret model behavior and complements leaderboard reporting with deeper context.

\section{Validating \method{}} \label{sec:validate}

In \S\ref{sec:applicability}, we previously demonstrated the utility of \method{} outcome; we now evaluate whether it functions as expected. The key contribution of this work is the resulting taxonomy. We present in this section multiple measures of it and attach an additional evaluation of the first stage (per-instance error analysis) and extra details in Appendix~\ref{appendix:eval}. The evaluations include mainly quantitative ones, conducted using \texttt{Qwen2.5-72B-Instruct} as a meta-judge, along with qualitative ones performed manually.

Building on prior work \citep{Wan2024TnTLLMTM, Shah2023UsingLL}, we evaluate the taxonomy using three main criteria: \emph{coverage}, how comprehensively the taxonomy captures error types; \emph{accuracy}, how reliably it categorizes them, and \emph{usefulness}, how well it aligns with the intended application. Usefulness is central to the taxonomy’s practical value, reflecting its support for downstream tasks; we therefore dedicate Section~\S\ref{sec:applicability} to this aspect. Additionally, we introduce robustness as a fourth criterion, motivated by concerns in the literature about the reliability of LLM-based evaluations \citep{10.1162/tacl_a_00681,siska-etal-2024-examining,lior2025reliableeval}. Robustness measures how stable the taxonomy remains under varying configuration setups.

\paragraph{Coverage}

The coverage is evaluated using the approach proposed by \citet{Wan2024TnTLLMTM}. For each sampled error instance, we attempt to automatically map it into the taxonomy. If no suitable category is found, we assign it to an ``other'' category. We further added ``hard to analyze'' category for cases where there is not enough information to analyze the example. As shown in Table~\ref{tab:error_atals_numbers}, only $1$ example was classified into the ``other'' category, and $48$ were classified into the ``hard to analyze''. We further include rare or uninformative categories, those not part of \product{}, as part of this group, totaling $295$ errors. We analyzed $7{,}049$ wrong predictions, and obtained a coverage score of \textbf{95.2\%}.

\paragraph{Accuracy}

We measure taxonomy accuracy using the approach of \citet{Wan2024TnTLLMTM}. In each evaluation round, we sample an error instance from the taxonomy and present it to a judge model. The judge is given the instance's assigned label from the taxonomy, along with an alternative negative label from it at random. Based on this information, the judge is asked to determine which label better fits the error. The results of this evaluation are shown in App. Table~\ref{tab:eval_results} and indicate a high level of agreement with the assigned labels, averaging an accuracy score of \textbf{92\%}.

\paragraph{Robustness}
We evaluated the robustness of the taxonomy construction process by manually comparing the original taxonomy against two configurations: an upper bound (larger sample and prompt variations) and a lower bound (smaller sample and smaller model). 
Overall, results suggest that our sampling strategy and approach are robust, and different models can capture common error patterns without significant gaps. More details can be found in Appendix~\ref{app:robust_eval_taxonomy}.



\section{Related Work}
While various works studied common errors in a specific setups, such as errors within a particular subdomain \citep{Dou2024WhatsWW,Wang2024TowardsUT,Ramprasad2024AnalyzingLB,deshpande2025trail} or errors specific to a single model \citep{yehudai2025clearerroranalysisllmasajudge}, or have defined challenges through the lens of question difficulty \citep{Bradley2024LLMsAT,baldock2021deep,hacohen2020let,choshen-etal-2022-grammar,habba-etal-2025-dove}, we are aware of no work that presents a general LLM error taxonomy or explores the use of global model error signals.
Although taxonomy construction can be difficult in domains where phrase meanings are ambiguous \cite{BabaeiGiglou2023LLMs4OLLL, Funk2023TowardsOC, ashury-tahan-etal-2024-data}, our setting is closer to data mining, where data is organized according to semantic relationships.
Moreover, taxonomies based on latent skills were used in contexts such as scaling laws \citep{polo2024sloth} and self-specialization \citep{Kang2023SelfSpecializationULA} to find, possibly non-interpretable, dimensions that describe shared model performance. \citet{maimon2025iqtestllmsevaluation} utilize latent skills for diagnosis, and create a static dedicated leaderboard to act as an IQ test for LLMs. Some manual efforts split input data to highlight instances posing a shared challenge \citep[e.g.,][]{Magnusson2023PalomaAB}, and others automate such practices \citep{choshen2019automatically,tjuatja2025behaviorboxautomateddiscoveryfinegrained}. Other works extract what each input tests to analyze model outputs \citep{zeng2025evaltreeprofilinglanguagemodel} or predict their skills \citep{zhou2025general,ruan2024observational,polo2024sloth}. We see great value in such works. While we explain the benefits of analyzing actual errors and relying on model outputs rather than the challenges in the inputs, we believe these works to be useful for different needs. 


We also note that a recurring part driving the decisions made in \method{} is efficiency. Currently, as it runs only on failed examples and batches it takes a similar amount of inference as the evaluation, in \product{} we also sample. A large body of works suggested ways of efficient sampling for evaluation, they also consistently find that for most cases a random sample is a strong baseline and most alternatives introduce a bias \citep{choshen-etal-2024-navigating,11123151,wang2025effieval,polotinybenchmarks,poloefficient,maia2024efficient,perlitz-etal-2024-efficient}. Since none of those methods aim for a distribution of errors and analysis, we sampled randomly.

\section{Conclusions}
In this work, we presented \method{}, a diagnosis method, to efficiently produce a summary of the errors models perform on a given benchmark, enabling more interpretable comparisons between models. We also introduced \product{}, a general taxonomy emphasizing current LLM errors, contributing to a deeper understanding of their behaviors. While obtaining a complete picture is an inherently challenging goal, \product{} offers a first glimpse into current model errors, based on a diverse set of benchmarks spanning multiple domains and skills. Our approach lays the foundation for more detailed model evaluation. We refer to the limitations of our work to App.~\ref{sec:limitations}.




\section*{Impact Statement}

We see diagnosis work as a net good. There are little obvious harms from better understanding where systems that we currently develop fail, except in aiding development of systems that shouldn't be developed in the same way it aids regular development. Importantly, our method supports responsible development, deployment, and auditing of models (see Table~\ref{tab:usecases}, particularly the domain expert persona).

\bibliography{example_paper}
\bibliographystyle{icml2026}

\newpage
\appendix
\onecolumn

\section{\method{}: Additional Details} \label{error_map_additional_analysis}\label{error_map_additional_details}

\begin{table}[h!]
\centering
\begin{tabular}{|l|c|p{6cm}|}
\hline
\textbf{Field Name} & \textbf{Default Value} & \textbf{Description} \\
\hline
\texttt{batch size} & 500 & Size of minibatches for data processing. \\
\texttt{classify batch size} & 50 & Size of minibatches for item classification. \\
\texttt{cluster name length} & 5 & Maximum length for cluster names. \\
\texttt{cluster description length} & 30 & Maximum length for cluster descriptions. \\
\texttt{max num of clusters} & 25 & Maximum number of clusters allowed. \\
\texttt{max layers} & 3 & Maximum number of category layers. \\
\hline
\end{tabular}
\caption{Taxonomy Configuration Parameters.}
\label{tab:taxonomy_config_params}
\end{table}

\subsection{Prompts}
\subsubsection{Per-Instance Error Analysis} \label{app:per_instance_prompt}
\begin{lstlisting}[frame=single, breaklines=true]

You are an expert analyst. Your job is to evaluate evidence step by step, consider alternatives, and reach a justified conclusion. Reasoning: high.

You are given the following:
- A context
- A model response that was labeled incorrect
{% if correct_answer %}
- A reference
{% endif %}
{% if correct_outputs %}
- A list of solutions that were labeled as correct
{% endif %}

Your task:

1. Structured Correct Solution: Analyze the correct responses and extract from them the main required criteria or reasoning steps for the context.

2. Step-by-step Evaluation: Evaluate the incorrect response against each of the required criteria. For each criterion, provide the following fields:
present_in_wrong: Whether it is present in the incorrect response
quality: The quality of its execution (correct, partially correct, incorrect, or null if missing)
evidence: Supporting evidence from the incorrect response (quote)
comment: Any relevant comments

3. Error Diagnosis: Identify the first major error in the incorrect response that led to the incorrect answer, and provide the following fields in final_answer:
error_summary: If such an error exists, summarize the model's reasoning weakness in error_summary. This should focus on model thinking (e.g., 'the model failed to recognize fact X') rather than technical execution (e.g., 'the model selected the wrong answer').
title: Provide a short, free-form title that describes the specific type of error.
* If you didn't find any error in the incorrect response leave all the fields of final_answer with an empty string.
* If the whole solution is incorrect, write 'whole solution incorrect' in final_answer fields.
* Avoid ambiguous titles or ones that cannot be mapped to a specific skill. For example, instead of using "Wrong multiple choice selection", identify the underlying reasoning error such as "Misinterpretation of concept".

Use as many steps and thinking process as you need. Finally, output the final result in the following format:
{
  "required_criteria": [
    {
      "criterion": "Describe the relationship between A and B",
      "present_in_wrong": true,
      "quality": "incorrect",
      "evidence": "Because A increased when B increased, A must be caused by B.",
      "comment": "Confuses correlation with causation"
    },
    {
      "criterion": "Explain the mechanism of action",
      "present_in_wrong": true,
      "quality": "correct",
      "evidence": "the biochemical pathway...",
      "comment": "Accurate and complete"
    }
  ],
  "final_answer": {
    "error_summary": "The incorrect response assumes causation from correlation, leading to a flawed conclusion about the relationship between A and B.",
    "error_title": "Causal Misinterpretation"
  }
}


Use the following inputs:

Context:
{{ input_text }}

{% if correct_answer %}
References:
{{ correct_answer }}
{% endif %}

{% if correct_outputs %}
Correct Responses:
{{ correct_outputs }}
{% endif %}

incorrect prediction:
{{ output_text }}

Keeping the evaluation criteria in mind, do not provide a general assessment. Be specific, structured, and evidence-based.

Assessment:
    
    \end{lstlisting}

\subsubsection{Taxonomy Generation Prompt} \label{app:taxonomy_generation_prompt}

\begin{lstlisting}[frame=single, breaklines=true]
You are an expert analyst. Your job is to evaluate evidence step by step, consider alternatives, and reach a justified conclusion. Reasoning: high.

# Instruction
## Context
- **Goal**: Your goal is to cluster the input data into meaningful categories for the given use case.
- **Data**: The input data will be a list of {{ data_type }} tuples, including the following elements:
    - **text**: {{ data_type }} as the first tuple element.
    - **num of occurrences**: number as the second tuple element.
- **Use case**: Generate a taxonomy that categorizes model errors based on the specific skills the model failed to demonstrate in each example.
{% if parent_category %}
All of these errors have already been labeled under the category: *{{ parent_category }}*.
Please ensure that you assign each error to more specific and informative sub-categories that go beyond the general label "{{ parent_category }}".
Focus on identifying the underlying skills or error types that provide deeper insight.
{% endif %}

## Requirements

#### Format

- Output clusters in **JSON format** with each cluster as an object in a `clusters` array.
- Each cluster must include:
  - `"id"`: category number starting from 1.
  - `"name"`: category name (max {{ cluster_name_length }} words), either a verb or noun phrase.
  - `"description"`: category description (max {{ cluster_description_length }} words).
- Example:
{
  "clusters": [
    {
      "id": 1,
      "name": "Factual Error",
      "description": "Model fails to retrieve or generate accurate information."
    }
  ]
}

- Total number of categories should be **no more than {{ max_num_clusters }}**.
- Output should be in **English** only.

### Quality

- **No overlap or contradiction** among the categories.
- **Name** is a concise and clear label for the category, identifies **one specific skill or ability only**. Use only phrases that are specific to each category and avoid those that are common to all categories.
- **Name** reflects core capabilities, not domain-specific contexts, or technical choices. 
  Example: not "Incorrect Anatomical Knowledge" but "Factual Error" (The issue is about factual accuracy, not biology specifically).
  If the issue does not clearly map to a specific skill, classify it as "Hard to Analyze" - this applies when the error is ambiguous, subjective, or lacks sufficient context to determine its nature confidently.
- **Description** differentiates one category from another.
- **Name** and **description** can **accurately** and **consistently** classify new data points **without ambiguity**.
- **Name** and **description** are *consistent with each other*.
- Output clusters match the data as closely as possible, without missing important categories or adding unnecessary ones.
- Output clusters should strive to be orthogonal, providing solid coverage of the target domain.
- Output clusters serve the given use case well.
- Output clusters should be specific and meaningful. Do not invent categories that are not in the data.


# Data
{{ data }}

# Questions
## Q1. Please generate a cluster list from the input data that meets the requirements.

Tips

- **User Feedback is MANDATORY**: You MUST address any previous user feedback in your clustering
- If user feedback was provided, explicitly explain how you've incorporated their specific concerns and suggestions
- The cluster list should be a **flat list** of **mutually exclusive** categories. Sort them based on their semantic relatedness.
- Though you should aim for {{ max_num_clusters }} categories, you can have *fewer than {{ max_num_clusters }} categories* in the cluster list;  but **do not exceed the limit.** 
- Be **specific** about each category. **Do not include vague categories** such as "Other", "General", "Unclear", "Miscellaneous" or "Undefined" in the cluster list.
- You can ignore low quality or ambiguous data points.

## Q2. Why did you cluster the data the way you did? Explain your reasoning **within {{ explanation_length }} words**. Include how you addressed any user feedback.

## Provide your answers in the tags: "clusters" - your generated cluster list with no more than {{ max_num_clusters }} categories, "explanation" - explanation of your reasoning process within {{ explanation_length }} words.

# Output

\end{lstlisting}

\subsubsection{Taxonomy Update Prompt} \label{app:taxonomy_update_prompt}
\begin{lstlisting}[frame=single, breaklines=true]
You are an expert analyst. Your job is to evaluate evidence step by step, consider alternatives, and reach a justified conclusion. Reasoning: high.

# Instruction
## Context
- **Goal**: You goal is to review the given reference list based on the input data for the specified use case, then update the reference list if needed.
    - You will be given a reference cluster list, which is built on existing data. The reference list will be used to classify new data points.
    - You will compare the input data with the reference list, output a rating score of the quality of the reference list, suggest potential edits, and update the reference list if needed.
- **Reference cluster list**: The input cluster list is in JSON format with each cluster as a "cluster" element, containing the following sub-elements:
    - **id**: category index.
    - **name**: category name.
    - **description**: category description used to classify data points.
- **Data**: The input data will be a list of {{ data_type }} tuples, including the following elements:
    - **text**: {{ data_type }} as the first tuple element.
    - **num of occurrences**: number as the second tuple element.
- **Use case**: Update the taxonomy that categorizes model errors based on the specific skills the model failed to demonstrate in each example.
{% if parent_category %}
All of these errors have already been labeled under the category: *{{ parent_category }}*.
Please ensure that you assign each error to more specific and informative sub-categories that go beyond the general label "{{ parent_category }}".
Focus on identifying the underlying skills or error types that provide deeper insight.
{% endif %}

## Requirements

### Format

- Output clusters in **JSON format** with each cluster as an object in a `clusters` array.
- Each cluster must include:
  - `"id"`: category number starting from 1.
  - `"name"`: category name (max {{ cluster_name_length }} words), either a verb or noun phrase.
  - `"description"`: category description (max {{ cluster_description_length }} words).
- Example:
{
  "clusters": [
    {
      "id": 1,
      "name": "Factual Error",
      "description": "Model fails to retrieve or generate accurate information."
    }
  ]
}

- Total number of categories should be **no more than {{ max_num_clusters }}**.
- Output should be in **English** only.

### Quality

- **No overlap or contradiction** among the categories.
- **Name** is a concise and clear label for the category, identifies **one specific skill or ability only**. Use only phrases that are specific to each category and avoid those that are common to all categories.
- **Name** reflects core capabilities, not domain-specific contexts, or technical choices. 
  Example: not "Incorrect Anatomical Knowledge" but "Factual Error" (The issue is about factual accuracy, not biology specifically).
  If the issue does not clearly map to a specific skill, classify it as "Hard to Analyze" - this applies when the error is ambiguous, subjective, or lacks sufficient context to determine its nature confidently.- **Description** differentiates one category from another.
- **Name** and **description** can **accurately** and **consistently** classify new data points **without ambiguity**.
- **Name** and **description** are *consistent with each other*.
- Output clusters match the data as closely as possible, without missing important categories or adding unnecessary ones.
- Output clusters should strive to be orthogonal, providing solid coverage of the target domain.
- Output clusters serve the given use case well.
- Output clusters should be specific and meaningful. Do not invent categories that are not in the data.

# Reference cluster list
{{ cluster_list }}

# Data
{{ data }}

# Questions
## Q1: Review the given reference list and the input data and provide a rating score of the reference list. The rating score should be an integer between 0 and 100, higher rating score means better quality. You should consider the following factors when rating the reference cluster list:
- **Intrinsic quality**:
    - 1) if the cluster list meets the *Requirements* section, with clear and consistent category names and descriptions, and no overlap or contradiction among the categories;
    - 2) if the categories in the cluster list are relevant to the the given use case;
    - 3) if the cluster list includes any vague categories such as "Other", "General", "Unclear", "Miscellaneous" or "Undefined".
- **Extrinsic quality**:
    - 1) if the cluster list can accurately and consistently classify the input data without ambiguity;
    - 2) if there are missing categories in the cluster list but appear in the input data;
    - 3) if there are unnecessary categories in the cluster list that do not appear in the input data.

## Q2: Explain your rating score in Q1 **within {{ explanation_length }} words**.

## Q3: Based on your review, decide if you need to edit the reference list to improve its quality. If yes, suggest potential edits **within {{ suggestion_length }} words**. If no, please output the original reference list.

Tips:
- You can edit the category name, description, or remove a category. You can also merge or add new categories if needed. Your edits should meet the *Requirements* section.
- The cluster list should be a **flat list** of **mutually exclusive** categories. Sort them based on their semantic relatedness.
- You can have *fewer than {{ max_num_clusters }} categories* in the cluster list, but **do not exceed the limit.**
- Be **specific** about each category. **Do not include vague categories** such as "Other", "General", "Unclear", "Miscellaneous" or "Undefined" in the cluster list.
- You can ignore low quality or ambiguous data points.

## Q4: If you decide to edit the reference list, please provide your updated reference list. If you decide not to edit the reference list, please output the original reference list.

## Provide your answers between the following tags:
"rating_score" - integer between 0 and 100
"explanation" - explanation of your rating score within {{ explanation_length }} words
"suggestions" - suggested edits within {{ suggestion_length }} words, or "N/A" if no edits needed
"clusters" - your updated cluster list if you decided to edit the reference list, or the original reference list if no edits made

# Output

\end{lstlisting}

\subsubsection{Taxonomy Review Prompt} \label{app:taxonomy_review_prompt}
\begin{lstlisting}[frame=single, breaklines=true]
You are an expert analyst. Your job is to evaluate evidence step by step, consider alternatives, and reach a justified conclusion. Reasoning: high.

# Instruction
## Context
- **Goal**: Your goal is to review the given reference list based on the requirements and the specified use case, then update the reference list if needed.
    - You will be given a reference cluster list, which is built on existing data. The reference list will be used to classify new data points.
    - You will compare the reference list with the requirements, output a rating score of the quality of the reference list, suggest potential edits, and update the reference list if needed.
- **Reference cluster list**: The input cluster list is in JSON format with each cluster as a "cluster" element, containing the following sub-elements:
    - **id**: category index.
    - **name**: category name.
    - **description**: category description used to classify data points.
- **Use case**: Review the taxonomy that categorizes model errors based on the specific skills the model failed to demonstrate in each example.
{% if parent_category %}
All of these errors have already been labeled under the category: *{{ parent_category }}*.
Please ensure that you assign each error to more specific and informative sub-categories that go beyond the general label "{{ parent_category }}".
Focus on identifying the underlying skills or error types that provide deeper insight.
{% endif %}

## Requirements

### Format

- Output clusters in **JSON format** with each cluster as an object in a `clusters` array.
- Each cluster must include:
  - `"id"`: category number starting from 1.
  - `"name"`: category name (max {{ cluster_name_length }} words), either a verb or noun phrase.
  - `"description"`: category description (max {{ cluster_description_length }} words).
- Example:
{
  "clusters": [
    {
      "id": 1,
      "name": "Factual Error",
      "description": "Model fails to retrieve or generate accurate information."
    }
  ]
}

- Total number of categories should be **no more than {{ max_num_clusters }}**.
- Output should be in **English** only.

### Quality

- **No overlap or contradiction** among the categories.
- **Name** is a concise and clear label for the category, identifies **one specific skill or ability only**. Use only phrases that are specific to each category and avoid those that are common to all categories.
- **Name** reflects core capabilities, not domain-specific contexts, or technical choices. 
  Example: not "Incorrect Anatomical Knowledge" but "Factual Error" (The issue is about factual accuracy, not biology specifically).
  If the issue does not clearly map to a specific skill, classify it as "Hard to Analyze" - this applies when the error is ambiguous, subjective, or lacks sufficient context to determine its nature confidently.- **Description** differentiates one category from another.
- **Name** and **description** can **accurately** and **consistently** classify new data points **without ambiguity**.
- **Name** and **description** are *consistent with each other*.
- Output clusters match the data as closely as possible, without missing important categories or adding unnecessary ones.
- Output clusters should strive to be orthogonal, providing solid coverage of the target domain.
- Output clusters serve the given use case well.
- Output clusters should be specific and meaningful. Do not invent categories that are not in the data.

# Reference cluster list
{{ cluster_list }}

# Questions
## Q1: Review the given reference list and provide a rating score. The rating score should be an integer between 0 and 100, higher rating score means better quality. You should consider the following factors when rating the reference cluster list:
    - **Intrinsic quality**:
        - If the cluster list meets the required quality with clear and consistent category names and descriptions, and no overlap or contradiction among the categories.
        - If the categories in the cluster list are relevant to the specified use case.
        - If the cluster list does not include any vague categories such as "Other", "General", "Unclear", "Miscellaneous" or "Undefined".
    - **Extrinsic quality**:
        - If the cluster list can accurately and consistently classify the input data without ambiguity.
        - If there are missing categories in the cluster list that appear in the input data.
        - If there are unnecessary categories in the cluster list that do not appear in the input data.

## Q2: Explain your rating score in Q1 [The explanation should be concise, based on the intrinsic and extrinsic qualities evaluated in Q1].

## Q3: Based on your review, decide if you need to edit the reference list to improve its quality. If yes, suggest potential edits [Suggestions should be specific, actionable, and within the constraints of the maximum number of categories and use case specificity].

## Q4: If you decide to edit the reference list, provide your updated reference list. If you decide not to edit the reference list, please output the original reference list.

## Provide your answers between the following tags:
"rating_score" - integer between 0 and 100
"explanation" - explanation of your rating score within {{ explanation_length }} words
"suggestions" - suggested edits within {{ suggestion_length }} words, or "N/A" if no edits needed
"clusters" - your updated cluster list if you decided to edit the reference list, or the original reference list if no edits made

# Output
\end{lstlisting}

\subsubsection{Error Label Classification Prompt} \label{app:classify_prompt}
\begin{lstlisting}[frame=single, breaklines=true]
You are an expert analyst. Your job is to evaluate evidence step by step, consider alternatives, and reach a justified conclusion. Reasoning: high.

Your task is to use the provided taxonomy to categorize the overall topic or intent of each error generated by LLMs.

First, here is the taxonomy to use:

{{ taxonomy }}

To complete the task:

1. Carefully read through the entire {{ data_type }}, which contains a list of errors.
2. For each error, consult the taxonomy and identify the **single most relevant category** that best captures the overall topic or intent of that specific error.
3. If no category fits well, use the category 'Other'.
4. Output the result in a JSON format, where each tuple contains the error text and its assigned category. Use the following format:


{
  "classified_errors": [
    {
      "error_text": "error text 1",
      "category": "category name 1"
    },
    ...
    ]
}


5. Do not assign multiple categories to a single error. Choose only one that best fits.
That's it! Think carefully and explain your reasoning before giving your final category choice for each error.

Assign a single category to each of the following errors:

{{ data }}

Respond with your categories within json format, one per error. Do not include the number, just the category text.

\end{lstlisting}

\subsection{Resulted Taxonomy Examples} \label{app:taxonomy_examples_3_layers}

\underline{Example Num. 1}

\textbf{Category:} Missing Required Element (Omits mandatory sections, fields, identifiers, or other specified content.)

\textbf{Sub-category} Missing Contextual Details (Key domain‑specific or contextual information is not provided.)

\textbf{Subsub-category:} Background Context Omission (Lacks overarching situational or background information necessary for understanding.)

\textbf{Label:} Missing Historical Basis

\textbf{Error Summary:}
The response omits the essential historical facts (marriage to Queen Maria and subsequent regency) that explain how John of Brienne actually acquired the title of King of Jerusalem, relying instead on vague inspirational statements.

\underline{Example Num. 2}

\textbf{Category:} Computation Error (Produces incorrect numerical, algebraic, or geometric results, including miscalculations and faulty derivations.)

\textbf{Sub-category} Formula Misapplication Error (Using an incorrect or inappropriate formula for the problem domain.)

\textbf{Subsub-category:} Spatial Geometry Error (Misuse of area, volume, perimeter, trigonometric, or other geometric relations.)

\textbf{Label:} Misapplication of General Position Formula

\textbf{Error Summary:}
The model misapplied the general‑position formula for the maximum number of regions created by n planes, assuming all nine planes are in general position, which they are not. This led to an inflated count of 166 instead of the correct 48.












\null\newpage

\section{Experimental Setup} 
\label{sec:exp_setup}

We begin by introducing the conducted experiments, followed by a description of the general configuration shared across them, and conclude with a summary of the compute resources used for each experiment.

We conducted three experiments, which provide examples for the flexible usage of our approach; (1) \product{} Construction (\S\ref{sec:atlas}, \S\ref{sec:results}): we sample from all selected data and models, (2) Model Comparison: we utilize the \product{} categories and run only stages 1 (Per-Instance Error Analysis) and 2.b (Error Assignment) 
in \method{} on all predictions of two Gemini models listed in the HELM Capabilities leaderboard.
We then present a comparative evaluation between them in Section \ref{subsec:model_developer}.
(3) Dataset Taxonomy: We demonstrate the application of \method{} to generate a taxonomy tailored to a specific dataset, MMLU-Pro benchmark, in Section \ref{subsec:benchmark_stakehold}.

\paragraph{Failure Threshold} \method{} relies on a distinction between failed and successful instances, in non-binary metrics we make this distinction through a threshold. For each benchmark, we rely on a single metric (the primary score in the benchmark if there are multiple), and define for each range of scores what is the threshold considered as error.\footnote{A higher threshold is preferred over a lower one, as it prevents false negatives, ensuring that genuine errors are not mistakenly excluded. False positives, on the other hand, are mostly filtered out during instance-level analysis.} For datasets evaluated with a binary score, the selection is straightforward. For others, we found that using an approximation of $0.7\%$ of the maximum score per instance yields good results.

\paragraph{Taxonomy Parameters} The error categorization had to be well-defined in each of its prompts to provide a specific output. As part of this stage, and following the approach described in \citet{Wan2024TnTLLMTM}, we defined a set of parameters tailored to our case, such as error label batch size, maximum length for category names, and others. The complete list of parameters and their corresponding values is provided in Table~\ref{tab:taxonomy_config_params}.

\paragraph{Selected Judge} All experiments were conducted using the \texttt{gpt-oss-120b} model \citep{openai2025gptoss120bgptoss20bmodel}, chosen for its scale and relevance to current state-of-the-art systems. To better leverage its strong reasoning capabilities, we adapted the prompts accordingly. We add maximum $3$ ICP (if any exist) to each prompt.

\paragraph{Compute} The required compute for \method{} depends on the number of incorrect predictions. \product{} creation required approximately $7{,}200$ inferences. Since most of these can run in parallel, the process took approximately $3$ hours. The Gemini model comparison required about $2{,}000$ inferences. The MMLU-Pro experiment required approximately $3{,}500$ inferences.

\paragraph{Reliability validation} We used the \texttt{sentence-transformers/all-MiniLM-L6-v2} model. Changing the embedder did not change results, maybe because our task goes beyond textual similarity and aims to capture the underlying skills implied by the labels.

\paragraph{Statistical Significance Test}
To assess whether the differences in model distributions are statistically significant, we conducted pairwise comparisons between models. Specifically, we used binomial probability tests to evaluate the likelihood that the observed performance of a weaker model could occur under the distribution of a stronger one.

\newpage
\section{\product{} Details and Results} \label{appendix:error_atlas_analysis}

\begin{table}[h]
    \centering
    \begin{tabular}{c|c}
        \textbf{Dataset Name} & \textbf{Benchmark} \\
        \hline
        MMLU-Pro & \multirow{5}{*}{HELM Capabilities} \\
        GPQA & \\
        OmniMATH & \\
        WildBench & \\
        IFEval & \\
        \hline
        ACIBench & \multirow{8}{*}{MedHELM} \\
        MedDialog (healthcare magic) & \\
        MedDialog (icliniq) & \\
        MEDEC & \\
        MediQA & \\
        MedicationQA & \\
        MTSamples procedures & \\
        MTSamples replicate & \\
        \hline
        Live Multiple & \multirow{14}{*}{BFCL} \\
        Live Parallel Multiple & \\
        Live Parallel & \\
        Live Simple & \\
        Multi Turn Base & \\
        Multi Turn Long Context & \\
        Multi Turn Miss Func & \\
        Multi Turn Miss Param & \\
        Multiple & \\
        Parallel Multiple & \\
        Parallel & \\
        Simple JAVA & \\
        Simple Javascript & \\
        Simple Python & \\
        \hline
        QTSumm & \multirow{4}{*}{ToRR} \\
        NumericNLG & \\
        SciGen & \\
        TableBench (data analysis) & \\
        \hline
        HumanEval & \multirow{4}{*}{\textemdash} \\
        HumanEval+ & \\
        MBPP & \\
        MBPP+ & \\

    \end{tabular}
    \caption{List of datasets used to create \product{}.}
    \label{tab:selected_datasets}
\end{table}

\begin{table}
\centering

\begin{tabular}{lrrr}
\toprule
Error Category & Num. of Datasets & Num. of Models & Prevalence (\%) \\
\midrule
Missing Required Element & 31 & 82 & 15.56 \\
Specification Misinterpretation & 27 & 70 & 11.5 \\
Logical Reasoning Error & 25 & 56 & 9.09 \\
Incorrect Identification & 21 & 56 & 8.98 \\
Computation Error & 22 & 54 & 8.45 \\
Output Formatting Error & 23 & 76 & 7.7 \\
Irrelevant/Extraneous Content & 24 & 53 & 7.06 \\
Factual Error & 22 & 49 & 5.72 \\
Incomplete Reasoning & 17 & 44 & 5.03 \\
Counting/Enumeration Error & 11 & 44 & 4.65 \\
Answer Selection Error & 16 & 37 & 4.39 \\
Naming/Symbol Error & 18 & 29 & 3.16 \\
Tool/API Usage Error & 16 & 19 & 2.83 \\
Unit Conversion Error & 12 & 23 & 1.82 \\
Inappropriate Refusal & 8 & 25 & 1.39 \\
False Positive Detection & 8 & 16 & 1.02 \\
Other & 4 & 15 & 0.8 \\
Error Detection Failure & 6 & 11 & 0.7 \\
Hard to Analyze & 3 & 3 & 0.16 \\
\bottomrule
\end{tabular}
\caption{\product{} Statistics: The original categories generated by \texttt{gpt-oss-120b}, with the number of models and datasets classified to each category, along with their prevalence percentages.
\label{tab:error_atals_numbers}}
\end{table}

\begin{table}[]
\centering
\begin{tabular}{lrr}
\toprule
error\_category & count & percentage \\
\midrule
Logical Reasoning Error & 947 & 35 \\
Computation Error & 613 & 22 \\
Incorrect Identification & 317 & 12 \\
Missing Required Element & 209 & 8 \\
Specification Misinterpretation & 173 & 6 \\
Counting/Enumeration Error & 148 & 5 \\
Answer Selection Error & 114 & 4 \\
Factual Error & 69 & 3 \\
Incomplete Reasoning & 91 & 3 \\
Unit Conversion Error & 34 & 1 \\
Error Detection Failure & 4 & 0 \\
Irrelevant/Extraneous Content & 10 & 0 \\
Naming/Symbol Error & 7 & 0 \\
Other & 4 & 0 \\
Output Formatting Error & 2 & 0 \\
Tool/API Usage Error & 2 & 0 \\
\bottomrule
\end{tabular}
\caption{MMLU-Pro, GPQA and Omni MATH error category statistics.}
\label{tab:mmlu_omni_gpqa_stats}
\end{table}

\begin{table}[htbp]

\begin{tabularx}{\textwidth}{|p{4cm}|p{5cm}|p{7cm}|}
\toprule
Depth 1 Category & Depth 2 Subcategory & Depth 3 Sub-subcategory \\
\midrule
Logical Reasoning Error & Misapplication of Rule & Legal/Regulatory Rule Misapplication; Inequality Misuse \\
 & Faulty Logical Inference & Incorrect Base/Inductive Step; Misinterpreted Problem Structure \\
Missing Required Element & Missing Output Generation & Answer‑Choice Mapping Failure; Empty or Irrelevant Output \\
 & Missing Prompt Repetition & Exact Repetition Omission \\
Computation Error & Floating‑Point Precision Errors & -- \\
 & Parameter Value Assignment Errors & Numeric Value Error; Parameter Omission \\
Incorrect Identification & Goal/Objective Identification Errors & -- \\
 & Role/Function Identification Errors & -- \\
Specification Misinterpretation & Clarification Request Misinterpretation & -- \\
 & Constraint Misinterpretation & Constraint Meaning Misinterpretation; Hard to Analyze \\
Output Formatting Error & Length and Count Constraint Violations & Character Count Violation; Formatting Structure Violation \\
 & Paragraph and Section Structure Errors & Document-level Structure Errors; Paragraph Boundary Errors \\
Irrelevant/Extraneous Content & Topic Irrelevance & Scope Overextension; Topic Misalignment \\
 & Redundant Verbosity & Overly Complex Language; Redundant Repetition \\
Counting/Enumeration Error & Aggregation Error & -- \\
 & Constraint Ignorance & Constraint Misapplication; Constraint Omission \\
Answer Selection Error & Answer Choice Formatting Error & -- \\
 & Answer Choice Mapping Error & Correct Reasoning, Wrong Choice; Result-to-Choice Mapping \\
Incomplete Reasoning & Absent Empirical Evidence & -- \\
 & Assumption Omission & -- \\
Factual Error & Unsupported Speculation & -- \\
 & Terminology Misuse & -- \\
Tool/API Usage Error & Missing Prerequisite Data Retrieval & -- \\
 & Missing or Wrong Directory Navigation & Incorrect Directory Path; Missing Directory Change/Creation \\
Naming/Symbol Error & Identifier Naming Mistakes & Contextual Identifier Misuse; Naming Syntax Violation \\
 & Argument Type or Value Mismatch & Argument Type Mismatch; Argument Value or Ordering Mismatch \\
Inappropriate Refusal & Failure to Refuse Disallowed & -- \\
 & Style/Tone Compliance Refusal & -- \\
Unit Conversion Error & Arithmetic/Formula Unit Error & -- \\
 & Inconsistent Unit Usage & -- \\
False Positive Detection & Improper Content Filter Trigger & -- \\
 & Wrong Entity Absence Assumption & -- \\
Error Detection Failure & Factual Error Detection & -- \\
 & Policy Violation Detection & -- \\
\bottomrule
\end{tabularx}
\caption{Hierarchy of error taxonomy: sampled examples of depth-2 and depth-3 per category in depth-1.}
\label{tab:taxonomy_hierarchy_samples_subcategories}
\end{table}

\newpage
\null
\subsection{Examples}

\subsubsection{Incomplete Response Examples} \label{appendix:incomplete_response_examples}

\begin{lstlisting}[frame=single, breaklines=true]
Context:\\ 
Solve the equation $a^3 + b^3 + c^3 = 2001$ in positive integers.

References:\\ 
$\boxed{\left[(10,10,1), (10,1,10), (1,10,10)\right]}$

incorrect prediction:
\\ We are asked to solve the equation $a^3 + b^3 + c^3 = 2001$ in positive integers.
\\ We can observe that $1^3 = 1$, $2^3 = 8$, $3^3 = 27$, $4^3 = 64$, $5^3 = 125$, $6^3 = 216$, $7^3 = 343$, $8^3 = 512$, $9^3 = 729$, $10^3 = 1000$, $11^3 = 1331$, $12^3 = 1728$, $13^3 = 2197$.
\\ Since $a, b, c$ are positive integers, we must have $a, b, c \le 12$.
\\ Also, we can assume without loss of generality that $a \ge b \ge c \ge 1$.
\\ If $a=12$, then $a^3 = 1728$, so $b^3 + c^3 = 2001 - 1728 = 273$.
\\ If $b=6$, then $b^3 = 216$, so $c^3 = 273 - 216 = 57$, which is not a perfect cube.
\\ If $b=5$, then $b^3 = 125$, so $c^3 = 273 - 125 = 148$, which is not a perfect cube.
\\ If $b=4$, then $b^3 = 64$, so $c^3 = 273 - 64 = 209$, which is not a perfect cube.
\\ If $b=3$, then $b^3 = 27$, so $c^3 = 273 - 27 = 246$, which is not a perfect cube.
\\ If $b=2$, then $b^3 = 8$, so $c^3 = 273 - 8 = 265$, which is not a perfect cube.
\\ If $b=1$, then $b^3 = 1$, so $c^3 = 273 - 1 = 272$, which is not a perfect cube.
\\ If $a=11$, then $a^3 = 1331$, so $b^3 + c^3 = 2001 - 1331 = 670$.
\\ If $b=8$, then $b^3 = 512$, so $c^3 = 670 - 512 = 158$, which is not a perfect cube.
\\ If $b=7$, then $b^3 = 343$, so $c^3 = 670 - 343 = 327$, which is not a perfect cube.
\\ If $b=6$, then $b^3 = 216$, so $c^3 = 670 - 216 = 454$, which is not a perfect cube.
\\ If $b=5$, then $b^3 = 125$, so $c^3 = 670 - 125 = 545$, which is not a perfect cube.
\\ If $b=4$, then $b^3 = 64$, so $c^3 = 670 - 64 = 606$, which is not a perfect cube.
\\ If $b=3$, then $b^3 = 27$, so $c^3 = 670 - 27 = 643$, which is not a perfect cube.
\\ If $b=2$, then $b^3 = 8$, so $c^3 = 670 - 8 = 662$, which is not a perfect cube.
\\ If $b=1$, then $b^3 = 1$, so $c^3 = 670 - 1 = 669$, which is not a perfect cube.
\\ If$a=10$, then $a^3 = 1000$, so $b^3 + c^3 = 2001 - 1000 = 1001$.
\\ If$b=10$, then $b^3 = 1000$, so $c^3 = 1001 - 1000 = 1$, so $c=1$.
Thus, $(10, 10, 1)$ is a solution.
\\ If$a=1$, $b=10$, $c=10$, then $1^3 + 10^3 + 10^3 = 1 + 1000 + 1000 = 2001$.
Thus, $(1, 10, 10)$ is a solution.
\\ If$a=10$, $b=1$, $c=10$, then $10^3 + 1^3 + 10^3 = 1000 + 1 + 1000 = 2001$.
Thus, $(10, 1, 10)$ is a solution.
\\ If $a=10$, $b=10$, $c=1$, then $10^3 + 10^3 + 1^3 = 1000 + 1000 + 1 = 2001$.
\\ Thus, $(10, 10, 1)$ is a solution.

Final Answer: 
\\ The final answer is $\boxed{(1, 10, 10)}$
% \end{lstlisting}

\begin{lstlisting}[language=json,caption={Sample JSON Output}]

{
  "required_criteria": [
    {
      "criterion": "Establish an upper bound for the variables (show that a, b, c $\leq$ 12).",
      "present_in_wrong": true,
      "quality": "correct",
      "evidence": "Since $a, b, c$ are positive integers, we must have $a, b, c \\le 12$.",
      "comment": "The bound is correctly derived from $13^3 > 2001$."
    },
    {
      "criterion": "Assume without loss of generality an ordering (e.g., a $\geq$ b $\geq$ c) to reduce duplicate work.",
      "present_in_wrong": true,
      "quality": "correct",
      "evidence": "Also, we can assume without loss of generality that $a \\ge b \\ge c \\ge 1$.",
      "comment": "Proper ordering is stated."
    },
    {
      "criterion": "Systematically examine all feasible values of a (starting from the maximum) and for each check whether the remaining sum can be expressed as two cubes.",
      "present_in_wrong": true,
      "quality": "partially correct",
      "evidence": "The response checks a $\eq$ 12, 11, 10 but stops there; it does not examine a $\eq$ 9, 8, $\ldots$ down to the lower bound.",
      "comment": "Missing cases means the search is incomplete; the criterion is only partially satisfied."
    },
    {
      "criterion": "Identify the concrete solution(s) when a = 10 (i.e., b = 10, c = 1).",
      "present_in_wrong": true,
      "quality": "correct",
      "evidence": "If $a=10$, then $b^3 + c^3 = 1001$ $\ldots$ $b=10$, $c=1$ $\ldots$ Thus, $(10,10,1)$ is a solution.",
      "comment": "Correctly finds the core solution."
    },
    {
      "criterion": "Account for all permutations of the found unordered triple to list every ordered solution.",
      "present_in_wrong": true,
      "quality": "partially correct",
      "evidence": "The answer mentions $(1,10,10)$ and $(10,1,10)$ but the final boxed answer only shows $(1,10,10)`. ",
      "comment": "Permutations are recognized, yet the final response lists only one of the three ordered triples."
    },
    {
      "criterion": "Conclude that the listed permutations are the only positive-integer solutions to 
      $a^3+b^3+c^3=2001$.",
      "present_in_wrong": false,
      "quality": "incorrect",
      "evidence": "",
      "comment": "The response never explicitly states that no other solutions exist."
    }
  ],
  "final_answer": {
    "error_summary": "The answer fails to present the complete set of solutions and does not demonstrate that no other triples exist; it only outputs one ordered triple while ignoring the other two permutations and omits a final uniqueness argument.",
    "error_title": "Incomplete solution set"
  }
}

Example_id: id3293
Model: google/gemini-2.0-flash-lite-preview-02-05
Dataset: omni_math

\end{lstlisting}

\begin{lstlisting}[frame=single, breaklines=true]    
Context:
what does tasmar contain

References:
ACTIVE INGREDIENT/ACTIVE MOIETY
Tolcapone (UNII: CIF6334OLY) 

INACTIVE INGREDIENTS
Lactose Monohydrate (UNII: EWQ57Q8I5X)
povidone K30 (UNII: U725QWY32X)
sodium starch glycolate type a potato (UNII: 5856J3G2A2)
talc (UNII: 7SEV7J4R1U)
magnesium stearate (UNII: 70097M6I30)
titanium dioxide (UNII: 15FIX9V2JP)
triacetin (UNII: XHX3C3X673)
sodium lauryl sulfate (UNII: 368GB5141J)
ferric oxide yellow (UNII: EX438O2MRT)
ferric oxide red (UNII: 1K09F3G675)
MICROCRYSTALLINE CELLULOSE (UNII: OP1R32D61U)
ANHYDROUS DIBASIC CALCIUM PHOSPHATE (UNII: L11K75P92J)
ETHYLCELLULOSE, UNSPECIFIED (UNII: 7Z8S9VYZ4B)


incorrect prediction:
Tasmar is a brand name for the drug tolcapone. Tolcapone is a medication used in the treatment of Parkinson's disease. It works as a catechol-O-methyltransferase (COMT) inhibitor, which helps to increase the levels and duration of action of levodopa, a common medication used to manage Parkinson's symptoms. Tasmar is typically used in combination with levodopa/carbidopa therapy to improve its effectiveness. It is important to note that Tasmar can have serious side effects, including liver toxicity, so its use is carefully monitored by healthcare providers.

\end{lstlisting}

\begin{lstlisting}[language=json,caption={Sample JSON Output}]
{
  "required_criteria": [
    {
      "criterion": "State the active ingredient of Tasmar",
      "present_in_wrong": true,
      "quality": "correct",
      "evidence": "Tasmar is a brand name for the drug tolcapone.",
      "comment": "The response correctly identifies Tolcapone as the active ingredient."
    },
    {
      "criterion": "List all inactive ingredients of Tasmar as given in the reference",
      "present_in_wrong": false,
      "quality": "incorrect",
      "evidence": "",
      "comment": "The response does not provide any of the inactive ingredients; it instead discusses clinical use and safety."
    }
  ],
  "final_answer": {
    "error_summary": "The model focused on the pharmacological description of Tasmar rather than enumerating its ingredient composition, omitting the required list of inactive ingredients.",
    "error_title": "Omission of Required Ingredient List"
  }
}

example_id: id553
model: openai/gpt-4o-2024-05-13
dataset: medhelm_v2_medication_qa
\end{lstlisting}

\subsubsection{Question Misinterpretation Examples}\label{appendix:ques_misinterpret_examples}

\begin{lstlisting}[frame=single, breaklines=true]    
Context:
This question refers to the following information.
"To slacken the tempo would mean falling behind. And those who fall behind get beaten. But we do not want to be beaten. No, we refuse to be beaten! One feature of the history of old Russia was the continual beatings she suffered because of her backwardness. She was beaten by the Mongol khans. She was beaten by the Turkish beys. She was beaten by the Swedish feudal lords. She was beaten by the Polish and Lithuanian gentry. She was beaten by the British and French capitalists. She was beaten by the Japanese barons. 
All beat her--because of her backwardness,
because of her military backwardness, cultural 
backwardness, political backwardness, industrial 
backwardness, agricultural backwardness. They beat her 
because it was profitable and could be done with impunity. 
You remember the words of the pre-revolutionary poet: "You are poor and abundant, mighty and impotent, Mother Russia." 
Those gentlemen were quite familiar with the verses of the old poet. They beat her, saying: "You are abundant," so one can enrich oneself at your expense. They beat her, saying: "You are poor and impotent," 
so you can be beaten and plundered with impunity. Such is the law of the exploiters--to beat the backward and the weak. It is the jungle law of capitalism. 
You are backward, you are weak--therefore you are wrong; 
hence you can be beaten and enslaved. You are mighty--
therefore you are right; hence we must be wary of you.
That is why we must no longer lag behind."
Joseph Stalin, speech delivered at the first All-Union Conference of Leading Personnel of Socialist Industry, February 4, 1931
The reference to the "continual beatings" most directly supports which of the following decisions in later Soviet foreign policy?

Candidate Answers:
A) The declaration of war against Japan at the end of World War II
B) The blockade of Berlin
C) The support for Cuban Revolution
D) The decision to build the Berlin Wall
E) The imposition of the Brezhnev Doctrine against anti-Soviet protesters in Europe
F) The signing of the Molotov-Ribbentrop Pact
G) The launch of Sputnik
H) The diplomatic recognition of the People's Republic of China
I) The invasion of Afghanistan
J) The creation of the Warsaw Pact

References:
J


incorrect prediction:
The passage emphasizes the historical vulnerability and backwardness of Russia, which led to exploitation and aggression by foreign powers. Stalin's speech highlights the need to avoid lagging behind in industrial and military strength to prevent future exploitation. This sentiment aligns with the Soviet Union's decision to assert its power and influence to avoid being perceived as weak or backward.

Let's evaluate the options:

- **A. The declaration of war against Japan at the end of World War II:** This decision was driven by Soviet strategic interests and the desire to regain territory, but it is not directly tied to the theme of avoiding backwardness.
- **B. The blockade of Berlin:** This was a Cold War strategy to assert Soviet power in Europe, but it does not directly relate to the historical theme of backwardness.
- **C. The support for Cuban Revolution:** Supporting revolutions abroad was part of Soviet efforts to expand influence, but it is not directly tied to the theme of backwardness.
- **D. The decision to build the Berlin Wall:** This was a measure to prevent East Germans from fleeing to the West, but it is not directly tied to the theme of backwardness.
- **E. The imposition of the Brezhnev Doctrine against anti-Soviet protesters in Europe:** This was a policy to maintain Soviet control over Eastern Europe, but it does not directly address the theme of backwardness.
- **F. The signing of the Molotov-Ribbentrop Pact:** This non-aggression pact with Nazi Germany was a pragmatic decision to buy time and avoid conflict, but it does not directly relate to the theme of backwardness.
- **G. The launch of Sputnik:** This was a demonstration of Soviet technological advancement and strength, directly addressing the need to avoid being perceived as backward.
- **H. The diplomatic recognition of the People's Republic of China:** This was a geopolitical move, but it does not directly relate to the theme of backwardness.
- **I. The invasion of Afghanistan:** This was an attempt to expand Soviet influence, but it does not directly address the theme of backwardness.
- **J. The creation of the Warsaw Pact:** This was a military alliance to counter NATO, asserting Soviet strength, but it is not directly tied to the theme of backwardness.

The **launch of Sputnik** (G) best aligns with the theme of Stalin's speech, as it demonstrated Soviet technological and industrial progress, countering the historical narrative of backwardness and vulnerability.

The correct answer is (G).
\end{lstlisting}

\begin{lstlisting}[language=json,caption={Sample JSON Output}]
{
  "required_criteria": [
    {
      "criterion": "Recognize that the passage stresses historical exploitation of Russia because of its backwardness and the need to avoid lagging behind militarily.",
      "present_in_wrong": true,
      "quality": "partially correct",
      "evidence": "The model notes: \"The passage emphasizes the historical vulnerability and backwardness of Russia, which led to exploitation and aggression by foreign powers.\"",
      "comment": "The model correctly identifies the theme of backwardness, but does not link it specifically to a security-oriented foreign-policy response."
    },
    {
      "criterion": "Link the theme of avoiding backwardness to a foreign-policy decision that provides collective security against external powers.",
      "present_in_wrong": true,
      "quality": "incorrect",
      "evidence": "The model argues that the launch of Sputnik \"directly addresses the need to avoid being perceived as backward.\"",
      "comment": "Sputnik is a technological showcase, not a collective security measure; the passage calls for defending against exploitation, not a prestige project."
    },
    {
      "criterion": "Exclude options that pertain to internal achievements or isolated diplomatic moves that do not create a defensive alliance.",
      "present_in_wrong": true,
      "quality": "incorrect",
      "evidence": "The model dismisses many options (e.g., Berlin Wall, Brezhnev Doctrine) as ``not directly tied to the theme of backwardness,'' yet incorrectly selects Sputnik, also an internal achievement.",
      "comment": "The exclusion reasoning is inconsistent; the correct exclusion should keep the Warsaw Pact and reject purely symbolic actions."
    },
    {
      "criterion": "Select the Warsaw Pact (option J) as the decision that most directly embodies the passage's call for collective defense against exploitation.",
      "present_in_wrong": false,
      "quality": "incorrect",
      "evidence": "The model never mentions the Warsaw Pact and instead chooses option G (Sputnik).",
      "comment": "Failure to consider the Warsaw Pact indicates the model missed the central security implication of the passage."
    }
  ],
  "final_answer": {
    "error_summary": "The model misinterpreted the passage's emphasis on historical exploitation and the need for collective military security, treating it as a call for technological prestige and consequently selecting Sputnik instead of the Warsaw Pact.",
    "error_title": "Misinterpretation of Historical Context Leading to Wrong Policy Choice"
  }
}
model: deepseek-ai/deepseek-v3
dataset: mmlu_pro_old
example_id: id5031
\end{lstlisting}

\newpage
\section{\method{} Applicability}

\subsection{Benchmark Stakeholders}\label{subsec:benchmark_stakehold}

\begin{figure}[h]
    \centering
    \includegraphics[width=0.5\linewidth]{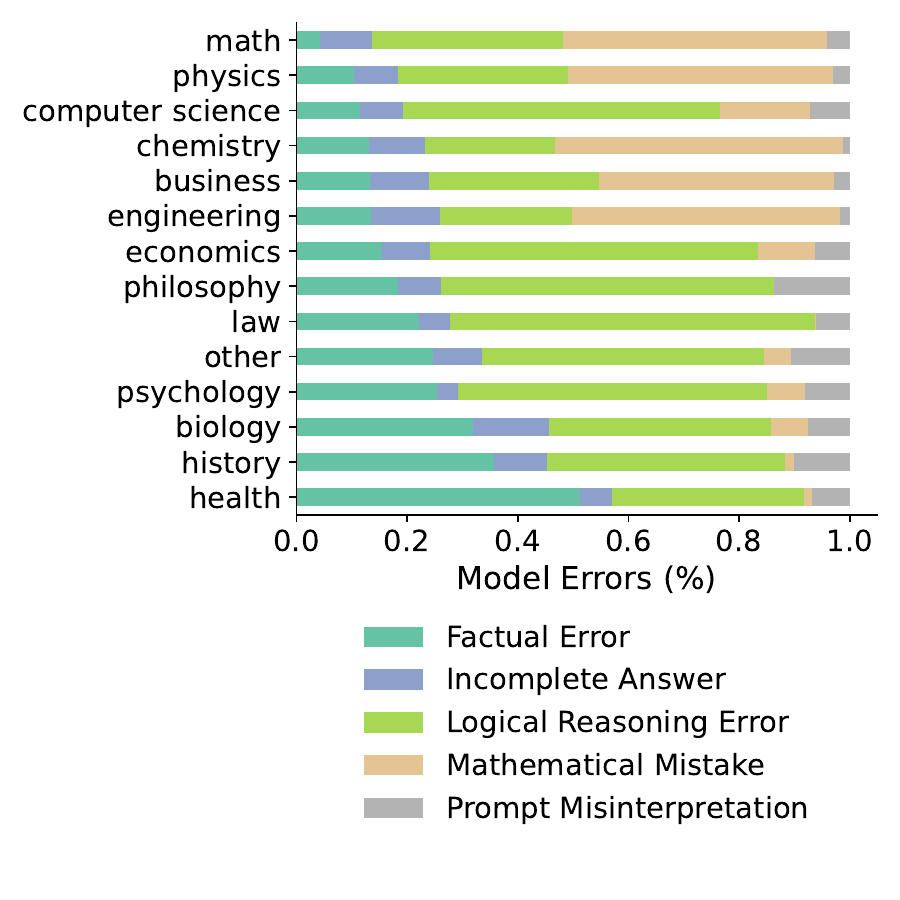}
    \caption{Differences in error category distributions across domain categories in MMLU-Pro.}
    \label{fig:mmlu_cat_vs_domains}
\end{figure}

\begin{table}[h]
    \footnotesize
    \centering
    \begin{tabular}{lr}
    \toprule
    Error Category & P-val ($\downarrow$) \\
    \midrule
    Factual Error & .000218 \\
    Incomplete Answer & .000000 \\
    Logical Reasoning Error & .000333 \\
    Mathematical Mistake & .002563 \\
    Prompt Misinterpretation & .074530 \\
    \bottomrule
    \end{tabular}
    \caption{Significance testing between the best- and worst-performing models for each error category. The results show that differences between models are usually statistically significant.}
    \label{tab:significance_best_worst}
    \vspace{-.5cm}
\end{table}

\section{\method{} Evaluation} \label{appendix:eval}

\subsection{Per-Instance Error Analysis}

Two key aspects of the per-instance analysis stage are accuracy and robustness. While accuracy measures whether the judge assigned a correct label to the error, robustness can be evaluated in multiple ways \citep{habba-etal-2025-dove}. In this work, we adopt a commonly used approach to measure robustness by examining the model’s sensitivity to prompt variations  \citep{Pezeshkpour2023LargeLM, Errica2024WhatDI, Zhuo2024ProSAAA}.

To evaluate \textit{accuracy}, we provided the meta-judge with all the information given to the original judge and its proposed analysis. Notably, the meta-judge’s task is significantly simpler than that of the analysis component. While the judge must generate a coherent explanation for the error, often requiring reasoning across multiple steps, the meta-judge only verifies whether the given analysis correctly explains the error. In other words, the judge performs a binary classification (correct/incorrect) based on a predefined context, without needing to produce or synthesize new information. 
The meta-judge accepted the vast majority of instances in each case with an average score of \textbf{91.1\%} (see Table~\ref{tab:eval_results}).

For assessing \textit{robustness}, we followed the approach of \citet{Kamoi2024EvaluatingLA}, executing the per-instance stage with $3$ prompt variations. We then compared the consistency of the $2$ resulting error analyses with the original prompt setup.
This comparison is challenging to automate, as the error labels are free-form and may differ in non-relevant ways (e.g., style or level of generality). To approximate a quantitative measure, we computed pairwise cosine similarity between error labels from the original and varied prompts using Sentence-BERT embeddings \citep{reimers2019sentencebertsentenceembeddingsusing}(details in App.\ref{sec:exp_setup}). The average similarity score was moderate \textbf{53\%}. we analyzed manually $100$ examples, and found them to be divided into the following categories:
$~45\%$ included the same underlying concept, but one phrase was partial to another in the text. Another $~30\%$ suffered from varying specificity in the labels and not disagreements. For instance, the labels ``missing temporal specification for midnight setting'' and ``missing explicit time reference'' received a similarity score of $0.21$. And $~25\%$ included a different error, and this may stem from the soft nature of errors, that may be called in multiple ways.

\begin{table}[h!]
    \centering
    \begin{tabular}{lcc}
    \toprule
       Stage & Experiment & Accuracy (\%) \\
    \midrule
       \multirow{3}{*}{\makecell{Per-Instance}} & \product{}  & $.89$ \\
                         & Gemini comparison & $.90$ \\
                         & MMLU-Pro taxonomy & $.94$ \\
       \multirow{3}{*}{\makecell{Taxonomy}} & \product{}  & $.92$ \\
                         & Gemini comparison & $.93$ \\
                         & MMLU-Pro taxonomy & $.91$ \\
    \bottomrule
    \end{tabular}
    \caption{\method{} and \product{} Evaluation results.}
    \label{tab:eval_results}
\end{table}

\subsection{Robustness Analysis For \product{}} \label{app:robust_eval_taxonomy}

To evaluate the robustness of the taxonomy construction stage, we used the high level categories aggregated in \product{}. We compared its high-level results created by \texttt{gpt-oss-120b} using $10\%$ of the data to two interesting cases:

\begin{enumerate}
    \item \textbf{Upper bound:} To test the robustness of our sampling approach, we ran the process with a rephrased prompt, different examples (a sample), and a larger sample of $15\%$.
    \item \textbf{Lower bound:} To test robustness with respect to the model and sample size, we attempted to create the taxonomy using \texttt{qwen-30b-thinking} with a smaller sample of $5\%$, to see how far the final result would deviate.
\end{enumerate}

The comparison was performed manually by examining the category descriptions and their resulting subcategories to ensure they were as similar as possible.

The upper bound shows a high correlation with the original result, achieving $88\%$ precision and $88\%$ recall. We further observed that the differences in categories occur mainly in those that appear less frequently in the data (as the original taxonomy is ordered by frequency) and are also less informative than others (e.g., empty response). Thus, even the missing categories belong to the long tail and not the core.

The lower bound demonstrates high precision and covers the majority of common categories ($8$ out of the top $10$). Its recall with respect to the original is moderate, which suggests that the amount of data may be critical when the sample size is a bit small for representing the error distribution.

In summary, our sampling choices appear reasonable, and different models can aggregate similar common errors without significant gaps. This strengthens our approach for identifying LLM error trends.

\begin{table}[h!]
\centering
\begin{tabular}{|p{5cm}|p{5cm}|p{5.4cm}|}
\hline
\textbf{Original Results} & \textbf{Results Config 1} & \textbf{Results Config 2} \\
\hline
Logical Reasoning Error & Logical Reasoning Error & Conceptual Misunderstanding, Causal Misinterpretation \\\hline
Missing Required Element & Missing Required Content, Policy Violation, Tone/Style Mismatch &  Overlooking Constraints, Failure to Follow Instructions, Constraint Violation \\
\hline
Computation Error & Numerical Computation Error, Syntax or Code Generation Error & Calculation Error, Incorrect Formula Application\\\hline
Incorrect Identification & Entity Misidentification  & Misidentification of Concept, Pattern Misidentification\\\hline
Specification Misinterpretation & Prompt or Task Misinterpretation & Misinterpretation of Problem, Data Misinterpretation, Causal Misinterpretation \\\hline
Output Formatting Error & Output Formatting Violation, Language Accuracy Error & Format Adherence Error, Boolean Syntax Error \\\hline
Irrelevant/Extraneous Content & Redundant or Extraneous Output, Irrelevant or Off‑Topic Generation & \\\hline
Counting/Enumeration Error & Statistical or Probability Miscalculation & \\\hline
Answer Selection Error & Answer Choice Selection Error & Incorrect Category or Label\\\hline
Incomplete Reasoning & Unjustified Assumption & Unverified Assumption, Incomplete Response \\\hline
Factual Error & Factual Inaccuracy & \\\hline
Tool/API Usage Error & Incorrect API/Parameter Usage, Edge‑Case Oversight & \\\hline
Naming/Symbol Error & Data Retrieval or Lookup Error & \\\hline
Inappropriate Refusal & Inappropriate Refusal & \\\hline
Unit Conversion Error & Incorrect Data Handling & Incorrect Data Handling\\\hline
False Positive Detection & & \\\hline
Error Detection Failure & & \\\hline
& Domain Knowledge Misapplication  & \\\hline
&  Empty Response &  \\

\hline
\end{tabular}
\caption{Comparative analysis of \product{} final error categories across different configurations as detailed in \S\protect\ref{sec:validate}.}
\label{tab:compare_atlas_with_perturb_refined}
\end{table}

\begin{table}[h!]
\centering
\begin{tabular}{|l|c|c|}
\hline
\textbf{Metric} & \textbf{Results Config 1 (\%)} & \textbf{Results Config 2 (\%)} \\
\hline
Precision & 0.88 & 1.00 \\
Recall    & 0.88 & 0.52 \\
F1-score  & 0.88 & 0.69 \\
\hline
\end{tabular}
\caption{Considering the original created taxonomy, we compared it to ones with other configurations (see Table~\protect\ref{tab:compare_atlas_with_perturb_refined}).}
\label{tab:robust_table_measures}
\end{table}

\newpage






\newpage

\section{Limitations}\label{sec:limitations}

\paragraph{Predictions Signal} While our method relies on model outputs, we acknowledge that the prediction signal can be partial compared to what actually happens inside the model (as opposed to white-box interpretability). 

\paragraph{Informative Prediction Dependence} \method{} focuses on predictions as the primary basis for analysis. This approach technically depends on informative predictions. If a model cannot be run in generative mode and does not explain its response (CoT is also acceptable), then its responses cannot be analyzed.

\paragraph{Error Category} We acknowledge that error categories are inherently soft, that is, a single mistake may reflect multiple underlying issues. For example, a model incorrectly stating that “mRNA carries amino acids to the ribosome” could indicate either a factual error or confusion about molecular roles.

\paragraph{\product{} Generality} While we have tried to create \product{} in the most varied way possible, there may be cases where it does not represent certain specific domains well.

\paragraph{LLM-based technique}
This work makes use of LLMs to analyze mistakes made by LLMs themselves. While this approach is somewhat circular, verification and comparison are generally easier than generation \citep{Simonds2025RLSRRL, Pang2023LanguageMS, Lin2024WildBenchBL}. However, this assumption may not always hold in practice.


\section{Usage in AI}
In this work, we used AI models exclusively for language-related tasks, such as rephrasing and surface-level linguistic transformations. It was further used for minor improvements to code style across the repo.


\end{document}